\documentclass[10pt,twocolumn,letterpaper]{article}

\usepackage[pagenumbers]{cvpr} 
\usepackage{cvpr}              

\usepackage[dvipsnames]{xcolor}

\definecolor{cvprblue}{rgb}{0.21,0.49,0.74}
\usepackage[pagebackref,breaklinks,colorlinks,citecolor=cvprblue]{hyperref}
\usepackage{booktabs}
\usepackage{algorithm}
\usepackage{algorithmic}
\usepackage{multirow}
\usepackage{gensymb}
\usepackage{makecell}
\usepackage{adjustbox}
\usepackage{threeparttable}
\usepackage{bbding}
\usepackage{makecell,pifont,multirow,multicol}
\usepackage{colortbl}
\usepackage[misc]{ifsym}

\def\paperID{1997} 
\def\confName{CVPR}
\def\confYear{2025}
\usepackage{makecell,pifont,multirow,multicol}

\usepackage{hyperref}
\usepackage{orcidlink}

\def\ModelName{RoboSense}

\def\logo{\makebox[38pt][l]{\raisebox{-1.0ex}{\includegraphics[height=30pt]{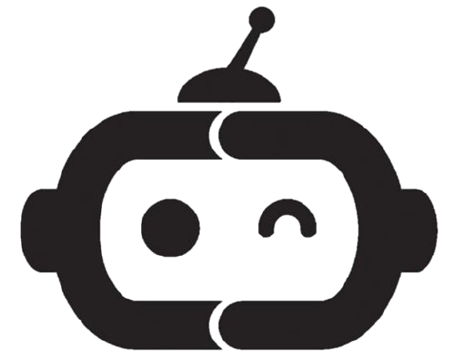}}}}

\title{\logo\ModelName: Large-scale Dataset and Benchmark for Egocentric Robot Perception and Navigation in Crowded and Unstructured Environments} 

\author{Haisheng Su$^{1,2}$ \quad
Feixiang Song$^{2}$ \quad
Cong Ma$^{2}$ \quad
Wei Wu$^{2}$ \quad
Junchi Yan$^{1}$$^{(\textrm{\Letter})}$ \\
$^{1}$School of Computer Science, Shanghai Jiao Tong University \\ $^{2}${SenseAuto Research} \\ 
{\tt\small \{suhaisheng,yanjunchi\}@sjtu.edu.cn, \{songfeixiang1,macong,wuwei\}@senseauto.com } \\
Code \& Dataset: \href{https://github.com/suhaisheng/RoboSense}{SenseAuto \& SJTU-ReThinkLab/RoboSense}
\vspace{-0.4cm}
}

\begin{document}
\maketitle

\begin{abstract}

Reliable embodied perception from an egocentric perspective is challenging yet essential for autonomous navigation technology of intelligent mobile agents. With the growing demand of social robotics, near-field scene understanding becomes an important research topic in the areas of egocentric perceptual tasks related to navigation in both crowded and unstructured environments. Due to the complexity of environmental conditions and difficulty of surrounding obstacles owing to truncation and occlusion, the perception capability under this circumstance is still inferior. To further enhance the intelligence of mobile robots, in this paper, we setup an egocentric multi-sensor data collection platform based on 3 main types of sensors (Camera, LiDAR and Fisheye), which supports flexible sensor configurations to enable dynamic sight of view from ego-perspective, capturing either near or farther areas. Meanwhile, a large-scale multimodal dataset is constructed, named RoboSense, to facilitate egocentric robot perception. Specifically, RoboSense contains more than 133K synchronized data with 1.4M 3D bounding box and IDs annotated in the full $360^{\circ}$ view, forming 216K trajectories across 7.6K temporal sequences. It has $270\times$ and $18\times$ as many annotations of surrounding obstacles within near ranges as the previous datasets collected for autonomous driving scenarios such as KITTI and nuScenes. Moreover, we define a novel matching criterion for near-field 3D perception and prediction metrics. Based on RoboSense, we formulate 6 popular tasks to facilitate the future research development, where the detailed analysis as well as benchmarks are also provided accordingly. Data desensitization measures have been conducted for privacy protection. 

\end{abstract}    

\begin{figure}[t]
\centering
\setlength{\abovecaptionskip}{-0.cm} 
\includegraphics[width=1.0\columnwidth]{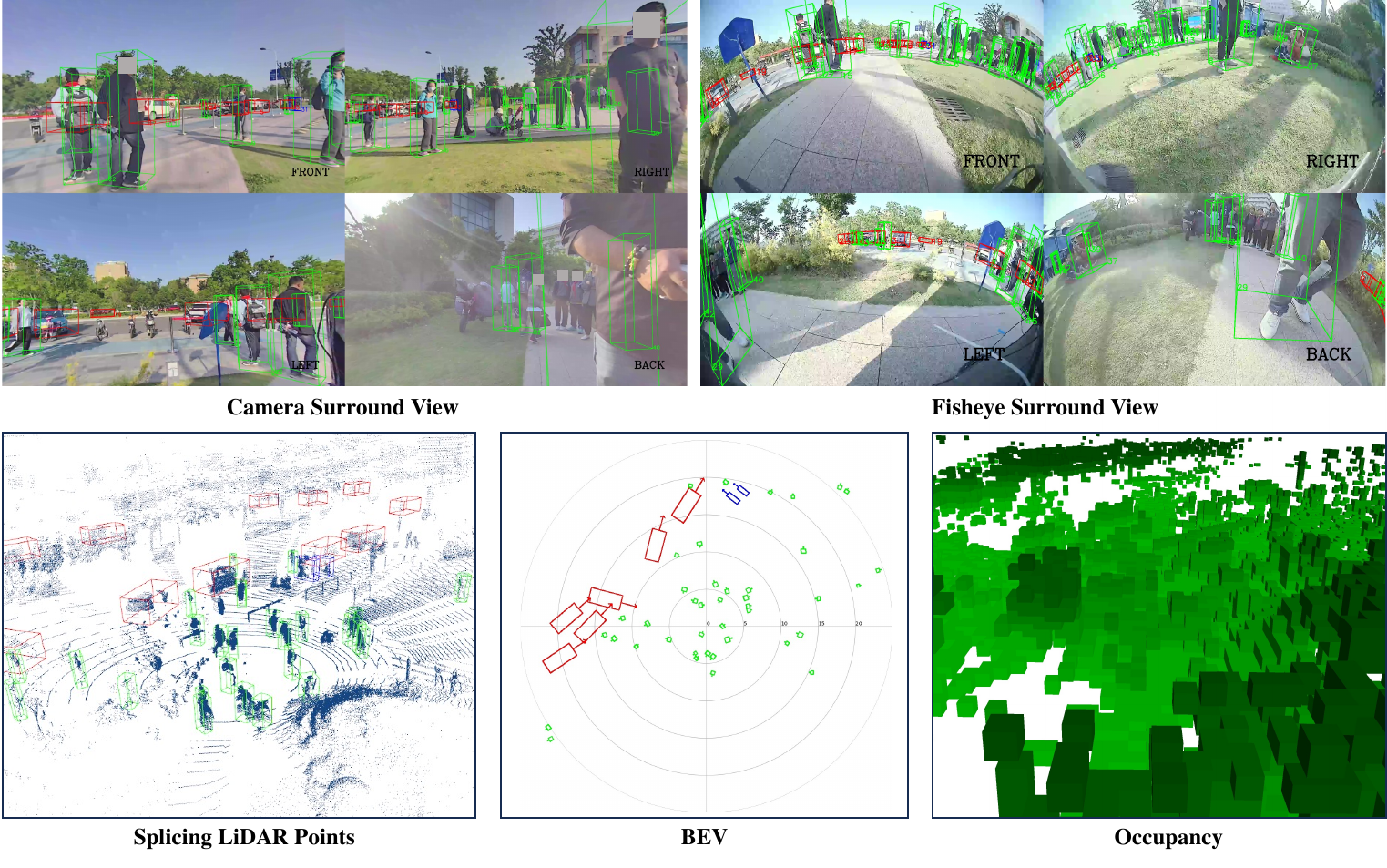}
\caption{An example from RoboSense dataset: The data with annotated 3D boxes and occupancy descriptions on Camera, Fisheye, LiDAR, and BEV respectively, where the same targets are associated with unique IDs across different devices and timestamps.}
\label{fig:dataset}
\vspace{-0.5cm}
\end{figure}

\vspace{-0.4cm}
\section{Introduction}
\label{sec:intro}

Recent years have witnessed significant progress achieved in the field of autonomous driving, enabling numerous intelligent vehicles running on highway or urban areas. In addition to self-driving cars, social mobile robots have emerged as a new industry tailored to autonomous navigation for typical applications, such as tractor, sweeper, retail and delivery. Notably, such intelligent mobile agents usually operate and navigate in crowded and unstructured environments (\textit{i.e.}, campuses, scenic spots, streets, parks and sidewalks, etc.), with varying and uncontrolled natural conditions such as illumination, occlusion and obstruction. In order to achieve navigation tasks safely, egocentric perceptual solutions enable these robots to perceive and comprehend the surrounding context from a first-person view, so as to interact successfully with passby pedestrians and vehicles, predict their intentions and incorporate this information in agents' planning and decision reasoning process.


To evaluate and compare different egocentric perceptual methods fairly, several standarized benchmarks~\cite{caesar2020nuscenes,geiger2012we,sun2020scalability,huang2018apolloscape,pham20203d,xiao2021pandaset} have been proposed in recent years, advancing the development of modern data-driven approaches. KITTI~\cite{geiger2012we} is a pioneering dataset providing multi-modal sensor data including front-view LiDAR pointclouds as well as corresponding stereo images and GPS / IMU data. nuScenes~\cite{caesar2020nuscenes} constructs a multi-sensor dataset collected in two cities travelling at an average of 16 km/h, where rich collections of 3D boxes and IDs are annotated in the full $360^{\circ}$ view. Waymo Open dataset~\cite{sun2020scalability} significantly increases the amount of annotations with higher annotation frequency. However, the target domain application of existing benchmarks is autonomous driving: the sensor data are captured exclusively from structural roads and highways, with sensor suites installed on top of cars.


To fill the vacancies of egocentric perceptual benchmarks target a unique domain related to navigation tasks in crowded and unstructured environments, in this paper, we present RoboSense, a novel multimodal dataset with several benchmarks associated to it. Our dataset is collected from diverse social scenarios filled with crowded obstructions, which is different from previously collected datasets used for autonomous driving (\textit{e.g.} nuScenes~\cite{caesar2020nuscenes}). Benefiting from the well time-synced multi-sensor data, we hope that our RoboSense can facilitate the development of egocentric perceptual frameworks for various types of autonomous navigation agents with controllable cost, not only self-driving cars but also autonomous agents such as social mobile robots. To this end, the data collection robot is equipped with 3 main types of sensors (C: Camera, L: LiDAR, F: Fisheye), and each type of sensor consists of 4 devices installed on different sides respectively to ensure the data captured under full $360^\circ$ view without blind spots.


Specifically, RoboSense consists of a total of 133K+ frames of synchronized data, spanning over 7.6K temporal sequences of 6 main scene classes (\textit{i.e.}, scenic spots, parks, squares, campuses, streets and sidewalks). Moreover, 1.4M 3D bounding boxes together with track IDs are annotated based on 3 different types of sensors, where most of targets tend to be closer to the robot as shown in Fig.~\ref{fig:dataset}. Then we form global trajectories for each agent separately through associating the same IDs across consecutive frames and different devices from a Bird's-Eye View (BEV) perspective. Additionally, we formulate 6 standarized benchmarks for egocentric perceptual tasks as follows: 1. Multi-view 3D Detection; 2. LiDAR 3D Detection; 3. Multi-modal 3D Detection; 4. Multiple 3D Object Tracking (3D MOT); 5. Motion Prediction; 6. Occupancy Prediction. Meanwhile, multi-task end-to-end training scheme is also supported in our RoboSense for evaluation of joint optimization. In sum, the main contributions of our work are three folds:


\begin{itemize}
\item To our best knowledge, our RoboSense is the first dataset tailored to egocentric perceptual tasks related to navigation of autonomous agents in unstructured environments.

\item We annotate 1.4M 3D bounding boxes on 133K+ synchronized sensor data, where most of targets are closer to the robot. Each target is associated with a unique ID, thus forming a total of 216K trajectories, which spread over 7.6K temporal sequences, covering 6 main scene classes.


\item We formulate 6 standardized benchmarks to facilitate the evaluation and fair comparisons of different perceptual solutions related to navigation in built environments.

\end{itemize}

\begin{table*}[t]\footnotesize
\begin{center}
\caption{Statistical comparison between RoboSense and similar existing datasets used for autonomous driving. C: Camera, L: LiDAR, F: Fisheye.  $\dagger$ means statistics exclude the testing set, which is unavailable. $\ddag$ indicates 10$\times$ higher annotation frequency (10Hz).} \label{table_dataset_comparison}
\vspace{-0.2cm} 
\setlength{\tabcolsep}{0.3cm}
		\begin{tabular}{l c c c c c c c c c}
			\toprule[1pt]
			\textbf{Dataset} & \textbf{Year} & \makecell{\textbf{Size} \\ \textbf{(hr)}} & \makecell{\textbf{Ann.} \\ \textbf{Scenes}}  & \makecell{\textbf{Ann.}\\ \textbf{Frames}} &\makecell{\textbf{With}\\ \textbf{Trajectory}} & \makecell{\textbf{Multi-view}\\ \textbf{Overlapping}} & \makecell{\textbf{Sensor} \\ \textbf{Layouts}} & \makecell{\textbf{3D Boxes} \\ \textbf{( Total )}} &
            \makecell{\textbf{3D Boxes$^{\dagger}$} \\ \textbf{( $\leq$ 5\textbf{\textit{m}} )}}
			\\
        \hline
        \hline
            KITTI~\cite{geiger2012we} & 2012 & 1.5  & 22 & 15K  & \ding{55} & \ding{55} & 4C+1L& 80K & 638 \\
        \hline
            Cityscapes~\cite{cordts2016cityscapes} & 2016 & -  & - & 25K  & \ding{55} & \ding{55} & 1C & 0 & 0 \\
        \hline
            ApolloScape~\cite{huang2018apolloscape} & 2016 & 2  & - & 144K  & \ding{55} & \ding{55} & 1L & 70K & 4.7K \\
        \hline
            H3D~\cite{patil2019h3d} & 2019 & 0.77  & 160 & 27K  & \ding{55} & \checkmark & 3C+1L & 1.1M & - \\
        \hline
            Lyft L5~\cite{Lyft5} & 2019 & 2.6  & 366 & 55K & \checkmark & \checkmark & 7C+3L & 1.3M & - \\
        \hline
            nuScenes~\cite{caesar2020nuscenes} & 2019 & 5.5  & 1K & 40K & \checkmark & \checkmark & 6C+1L & 1.4M & 9.8K \\
       \hline
            Argoverse~\cite{chang2019argoverse} & 2019 & 0.6  & 113 & 22K & \checkmark & \checkmark & 9C+2L & 993K & 15K \\
        \hline
            Waymo Open~\cite{sun2020scalability}  & 2019 & 6.4  & 1K  & 200K$\ddag$ & \checkmark & \checkmark & 5C+5L & 12M$\ddag$ & 123K$\ddag$  \\
        \hline
            BDD100k~\cite{yu2020bdd100k} & 2020 & 1K  & 100k & 100k  & \ding{55} & \ding{55} & 1C & 0 & 0 \\
        \hline
        \hline
            RoboSense (\textbf{Ours}) & 2024 & 42  & 7.6K & 133K & \checkmark & \checkmark & 4C+4F+4L & 1.4M & 173K \\
        \bottomrule[1pt]
\end{tabular}
\end{center}
\vspace{-0.5cm} 
\end{table*}

\section{Related Work}
We summarize the compositions of some existing perception and prediction datasets as shown in Tab.~\ref{table_dataset_comparison}.

\noindent
\textbf{Perception Datasets.} Current released perception datasets can be divided into image-only datasets~\cite{yu2020bdd100k,cordts2016cityscapes} and multimodal datasets~\cite{geiger2012we,caesar2020nuscenes,sun2020scalability,Lyft5,huang2018apolloscape}. BDD100k~\cite{yu2020bdd100k} and Cityscapes~\cite{cordts2016cityscapes} focus on 2D perception which provide large amount of 2D annotations (boxes, masks) for driving scene understanding under various weather and illumination conditions. KITTI~\cite{geiger2012we} is known as the pioneering multimodal dataset which has been widely used for academic research. It records 6 hours of driving data using a LiDAR sensor and a front-facing stereo camera to provide pointclouds and images with annotated 3D boxes. H3D dataset~\cite{patil2019h3d} collects a total of 1.3M 3D objects over 27K frames from 160 crowded scenes of the full 360$^\circ$ view. nuScenes~\cite{caesar2020nuscenes} and Waymo Open Dataset~\cite{sun2020scalability} are two similar datasets with same structure, while the latter one providing more annotations owing to higher annotation frequency (2Hz vs. 10Hz). \textit{Different from previously collected datasets used for autonomous driving, the annotation frequency of our RoboSense is even smaller (1Hz) due to the low speed (less than 1 m/s) moving status of social mobile robots navigating in crowded and unstructured environments.}


\noindent
\textbf{Prediction Datasets.} nuScenes~\cite{caesar2020nuscenes} and Waymo Open Dataset~\cite{sun2020scalability} can be also used for prediction task which release lane graphs as well. Lyft~\cite{Lyft5} introduces traffic/speed control data, and Waymo Open Dataset~\cite{sun2020scalability} adds more signals to the map such as crosswalk, lane boundaries, stop signs and speed limits. Recently, Shifts dataset~\cite{malinin2021shifts} becomes the largest forecasting dataset with the most scenario hours to date. Meanwhile, Argoverse~\cite{chang2019argoverse} is also a large-scale dataset with high data frequency (10Hz) and high scenario quality for motion forecasting ($> 2000 km$ across 6 cities). Together, these datasets have enabled exploration of multi-actor, long-range motion forecasting leveraging both static and dynamic maps.

\begin{figure}[t]
\centering
\setlength{\abovecaptionskip}{-0.cm} 
\includegraphics[width=1.0\columnwidth]{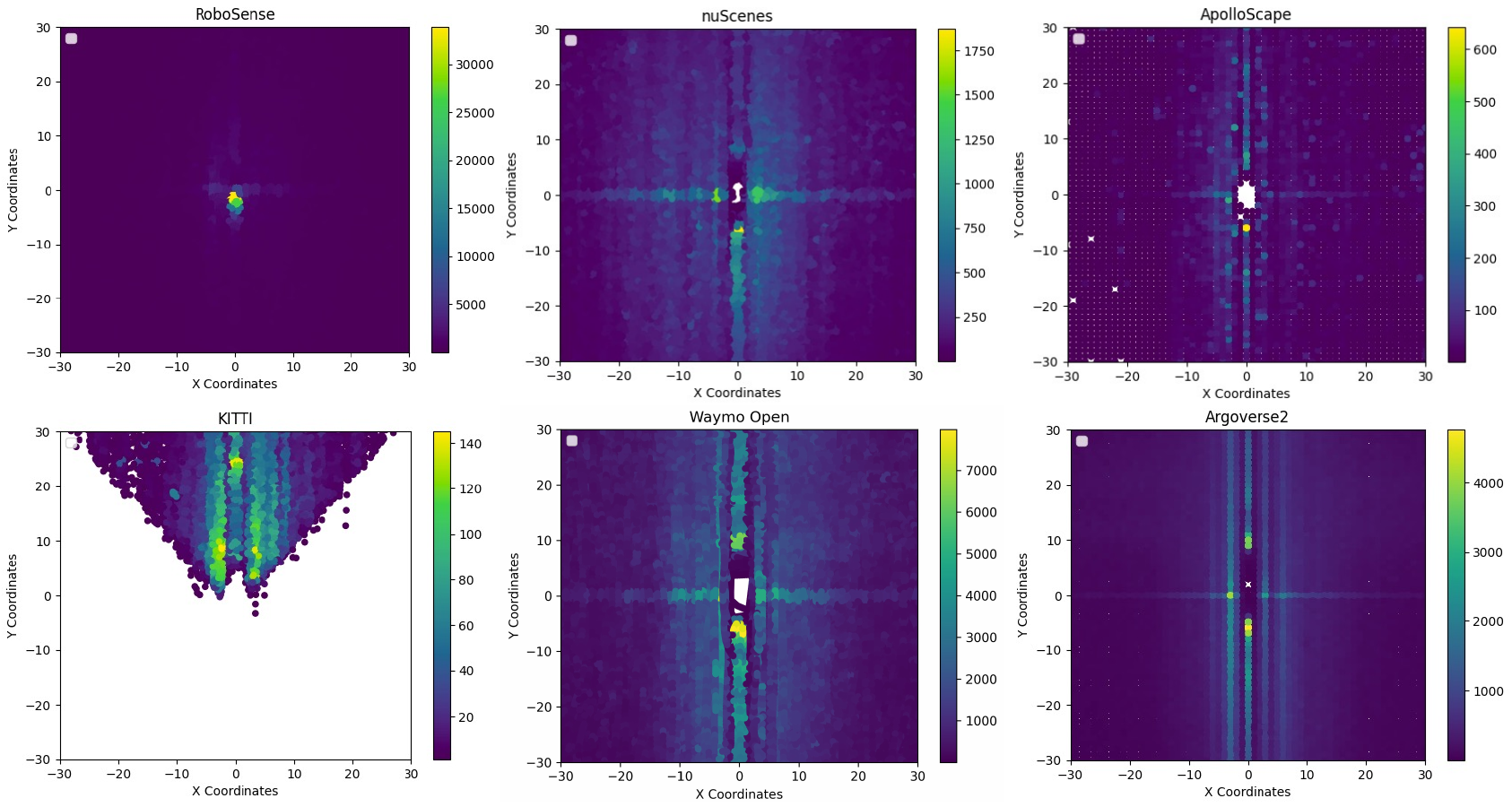}
\caption{Comparison of annotated object distribution among different popular datasets used for perception and prediction tasks.}
\label{fig:dataset_comparison}
\vspace{-0.3cm}
\end{figure}

\textbf{\textit{Generally, our dataset differs in three substantial ways}}: 1) targets a unique domain related to navigation tasks in crowded and unstructured environments, which is more difficult than autonomous driving scenarios in terms of complexity of environmental context and diversity of surrounding obstructions. 2) In addition to 3D bounding box and trajectory annotations, our dataset also provides high-quality occupancy descriptions for each collected scene, supporting the occupancy prediction task around the social robotics for safe navigation. 3) Our dataset is mostly collected in social crowded scenes, where pedestrians and cars tend to be closer to the robot, yielding a distribution with a mode at approximately 5$m$, which is quite different to the existing datasets for autonomous cars as shown in Fig.~\ref{fig:dataset_comparison}. Besides, the egocentric perceptual tasks under this circumstance is more challenging due to frequent occlusion and truncation.


\section{RoboSense Open Dataset}


We commence with the sensor setup as well as data acquisition details, delineate the coordinate systems and label generation process, and present data statistics respectively.

\subsection{Sensor Setup and Data Acquisition}
\begin{figure}[t]
\centering
\setlength{\abovecaptionskip}{-0.cm} 
\includegraphics[width=1.0\columnwidth]{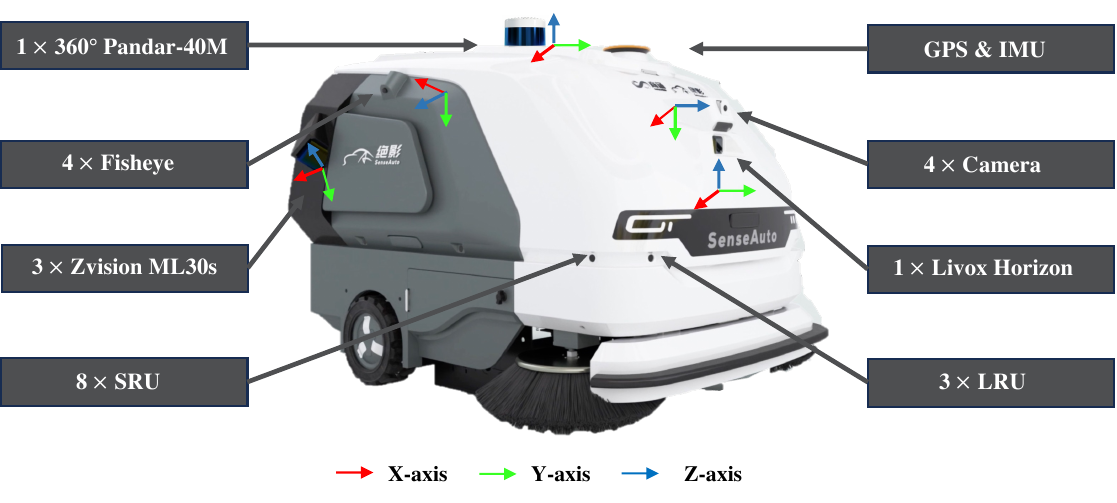}
\caption{Sensor setup and coordinate system illustration of our data collection platform.}
\label{fig:platform_setup}
\end{figure}

\noindent
\textbf{Sensor setup.} We use a social mobile robot (\textit{i.e.}, robosweeper) as data collection platform, which is equipped with different sensors installed in different sides of the robot respectively to ensure data captured in 360$\degree$ horizontal view without blind spots, including LiDAR, Camera, Fisheye, GPS / IMU and Ultrasonic. Refer to Fig.~\ref{fig:platform_setup} for sensor layouts and Tab.~\ref{tab_sensor_specification} for detailed sensor specifications.

\noindent
\textbf{Data acquisition.} We utilize the mobile robot to collect data along the Dishui Lake in Shanghai, China, lasting 42h in total at an average speed of less than 1m/s through manually remote control. 22 different places are travelled, which can be categorized into 6 main kinds of outdoor or semi-closed social scenarios (\textit{i.e.}, scenic spots, parks, squares, campuses, streets and sidewalks). After data collection, we manually select and process 7619 representative scenes of 20$s$ duration respectively for further annotation, \textit{covering various natural conditions (\textit{i.e.}, weather and illumination) and diverse environmental background and obstructions (\textit{i.e.}, motion, amount, type, occlusion, truncation).}


\subsection{Coordinate Systems}

\noindent
\textbf{Ego-Vehicle Coordinate.}
The Ego-Vehicle Coordinate System is centered at the rear axle of the vehicle. The positive directions of the X, Y, and Z axes correspond to the forward, leftward, and upward directions of the vehicle, respectively. Ego-Vehicle Coordinate System is the most frequently used in tasks such as perception, tracking, prediction, and planning, where dynamic and static targets as well as trajectories are transformed into this coordinate system.

\noindent
\textbf{Global Coordinate.}
To transform the dynamic and static elements from historical and future frames into the current frame coordinate system, we need to establish a global coordinate system to record the position and orientation of the ego vehicle in each frame. The origin of the Global Coordinate System is an arbitrarily defined point in Shanghai Lingang, China, and the positive directions of the X, Y, and Z axes follow the definition of the North-East-Up coordinate.

\noindent
\textbf{LiDAR Coordinate.}
The LiDAR Coordinate System is defined based on the Hesai lidar installed directly above the vehicle, the positive directions of the X, Y, and Z axes follow the definition of the Ego-Vehicle Coordinate System.

\noindent
\textbf{Camera Coordinate.}
The RoboSweeper is equipped with four fisheye cameras and four pinhole cameras. The origin of the Camera Coordinate System for both types of cameras is the optical center. However, the positive directions of the coordinate axes are defined differently in the RoboSense dataset. In the fisheye coordinate system, the X, Y, and Z axes correspond to directly below, right, and behind the optical center, respectively. In contrast, in the pinhole coordinate system, these axes correspond to directly right, below, and front of the optical center, respectively.

\noindent
\textbf{Pixel Coordinate.}
The image is presented in the form of pixels, each pixel corresponds to a 2D pixel coordinate. The origin of the Pixel Coordinate System is the upper left corner of the image. Points in the 3D Camera Coordinate System can obtain coordinates in the Pixel Coordinate System through the camera projection.

\subsection{Ground Truth Labels}
After integrating, synchronizing and calibrating the multi-sensor raw data, we annotate keyframes (LiDAR, image) at the frequency of 1Hz due to the low-speed moving status.



\noindent
\textbf{3D object.} With the selected scenes of collected RoboSense dataset, we annotate 3D object boxes of 3 movable classes (\textit{i.e.}, ``Vehicle", ``Cyclist" and ``Pedestrian") for each sampled keyframe in both the LiDAR coordinate of pointclouds and the Camera coordinate of multi-view images respectively. Each annotated 3D box can be represented as $[x, y, z, w, l, h, \theta, cls]$, where $x, y, z$ indicate the 3D position of a regular object, and $w, l, h$ represent the scale information including width, length and height. $\theta$ and $cls$ correspond to the orientation (especially yaw angle) and the object class respectively. A three-stage auto-labelling pipeline is detailed in the supplementary material (see Sec.~\ref{sec:label_gen}).


\noindent
\textbf{Trajectory.} To facilitate the temporal tasks such as multi-object tracking and motion forecasting described in Sec.~\ref{sec.4}, we assign a unique Track ID $\tau$ to each agent across a temporal sequence on Bird-Eye-View (BEV) of the Ego-Vehicle coordinate. Furthermore, agents with the same $\tau$ within a sequence are linked together to form object trajectories.


\noindent
\textbf{Occupancy label.} In addition to 3 typical classes of moving objects on roads which are annotated temporally as above, there also exists a rich collection of static obstacles with irregular shapes especially in the complex scenarios (\textit{i.e.}, parks, campuses and squares, etc.) of RoboSense. To detailly describe the environment in surrounding camera views for driving safety, we voxelize the 3D space and generate high-quality yet dense occupancy labels to represent the voxel states. Similar with previous occupancy benchmarks~\cite{tong2023scene, tian2024occ3d} built upon public datasets~\cite{caesar2020nuscenes, sun2020scalability}, we conduct dynamic objects and static scenes segmentation along the temporal dimension based on annotated 3D boxes and trajectories. Then sparse LiDAR points inside each box are extracted from $T-k$ to $T+k$ frames respectively, where $T$ indicates the index of current keyframe, and $k$ is set to 10 empirically. Refer to the supplementary material for more details of occupancy label generation process (see Sec.~\ref{sec:occ_gen}).





\section{Tasks \& Metrics}
\label{sec.4}

Both egocentric perceptual tasks and prediction tasks are supported in our RoboSense dataset and benchmark.



\subsection{Perception}
\subsubsection{3D Object Detection}

The RoboSense 3D detection task requires to detect 3D bounding boxes of three main classes (i.e. ``Vehicle", ``Pedestrian" and ``Cyclist"), including position, size, orientation and category. Following the conventions in~\cite{geiger2013vision,caesar2020nuscenes,sun2020scalability}, we adopt mAP (mean Average Precision), AOS (Average Orientation Similarity) and ASE (Average Scale Error) to measure the performance of different detectors.

There are several matching criteria to define the true positive for Average Precision (AP) metric calculation. For example,~\cite{geiger2013vision} adopts 3D Intersection-over-Union (IoU) to match each prediction with a ground-truth box, while~\cite{caesar2020nuscenes} define a match through thresholding the 2D center distance on the Bird-Eye-View ground plane. As for RoboSense detection task, we also adopt a similar distance measure. Differently, we define the threshold as a relative $Proportion$ $p$ of ground truth \textit{Closest Collision-point Distance} (CCDP) from the ego-vehicle, rather than an absolute \textit{Center Distance} (CD) $d$ adopted in~\cite{caesar2020nuscenes}. We claim that the localization accuracy of near obstacles' \textit{closest collision-point} is more important in low-speed driving scenarios. Then AP is calculated as the normalized area under precision-recall curve~\cite{everingham2010pascal}. Finally, mAP is obtained by averaging over all classes $\mathbb{C}$ and matching thresholds $\mathbb{P}=\{5\%, 10\%, 20\%\}$:
\begin{equation}
  mAP = \frac{1}{|\mathbb{C}|\cdot{|\mathbb{P}}|}\sum_{c\in \mathbb{C}}\sum_{p\in\mathbb{P}}{AP}_{c,p}
\end{equation}

In addition to AP, we also measure AOS and ASE for each matched true positive, which represent the precision of predicted yaw angle and object scale respectively.  AOS (Average Orientation Similarity) is formulated as:
\begin{equation}
  AOS = \frac{1}{|\mathcal{R}|}\sum_{r\in \mathcal{R}}\mathop{\operatorname{max}}\limits_{\widetilde{r}:\widetilde{r}\geq r}{s}(\widetilde{r}),
\end{equation}
\begin{equation}
  s(r)=\frac{1}{|\mathbb{D}(r)|}\sum_{i\in \mathbb{D}(r)}\frac{1+cos\Delta_{\theta}^{(i)}}
  {2},
\end{equation}
where $\mathcal{R}$ indicates the recall range $[0.1,1]$ interpolated with 40 points. $\mathbb{D}(r)$ indicates the set of matched true positives at recall $r$. And $\Delta_{\theta}^{(i)}$ denotes the angle difference between sample $i$ and ground truth. Different from~\cite{sun2020scalability}, we only consider true positive samples under each recall level, rather than all predicted positives.

ASE is defined as $\mathbf{1-IoU}$, which aims to measure the scale error through calculating the 3D $\mathbf{IoU}$ after aligning orientation and translation of predictions with ground truth.

\subsubsection{Multi-Object Tracking} The tracking task is designed to associate all detected 3D boxes of movable object classes across input multi-view temporal sequences (i.e. videos or point cloud sequences). Each object is assigned a unique and consistent track ID $\tau$ from first appearance until complete vanishing. As for performance evaluation, we refer to~\cite{caesar2020nuscenes,geiger2013vision,luo2021multiple,sun2020transtrack}, and mainly adopt sAMOTA (Scaled Average Multi-Object Tracking Accuracy), AMOTP (Average Multi-Object Tracking Precision) to measure the 3D tracking performance. 

Formally, sAMOTA is defined as the mean value of sMOTA over all recalls:
\begin{equation}
  sAMOTA=\frac{1}{|\mathcal{R}|}\sum_{r\in \mathcal{R}}{sMOTA}_{r},
\end{equation}
{\fontsize{7.5}{1}\selectfont
\begin{equation}
  sMOTA_{r}=max(0, 1-\frac{FP_{r}+FN_{r}+IDS_{r}-(1-r)\cdot GT}{r\cdot GT}),
\end{equation}
}
where $FP_{r}, FN_{r}$ and $IDS_{r}$ represent the number of false positives (wrongly detection), false negatives (missing detection) and identity switches at the corresponding recall $r$, respectively. Similarly, AMOTP is the average results of MOTP among different recalls, which can be defined as:
\begin{equation}
  AMOTP=\frac{1}{|\mathcal{R}|}\sum_{r\in \mathcal{R}}\frac{\sum_{i,t}d_{i, t}}{TP_{r}},
\end{equation}
where $TP_{r}$ is the number of true positives at the recall $r$, and $d_{i, t}$ denotes the position error of matched track $i$ at timestamp $t$. Besides, additional metrics such as MT (Most Tracked) and ML (Most Lost)~\cite{bernardin2008evaluating} are also reported.


\subsection{Prediction}

\subsubsection{Motion Forecasting} Based on perception results, the motion forecasting task requires to predict agents' future trajectories. Specifically, $\mathcal{K}$ plausible trajectories in future $T=3s$ timesteps for each agent are forecasted as offsets to the current agent's position. Following the standard protocols~\cite{liang2020pnpnet,luo2018fast,peri2022forecasting,gu2023vip3d}, we adopt minADE (minimum Average Displacement Error), minFDE (minimum Final Displacement Error), MR (Miss Rate) and EPA (End-to-end Prediction Accuracy) as metrics to measure the precision of motion prediction. In order to decouple the accuracy of perception and prediction, these metrics are only caculated for matched TPs (True Positives), where the matching threshold is set to $p_{match}=5\%$ of ground truth distance of the closest collision-point from the ego-vehicle. And the miss threshold of minFDE is set to $p_{miss}=20\%$ for calculating the MR metric.

\subsubsection{Occupancy Prediction} The goal of occupancy prediction task is to estimate the state of each voxel in the 3D space. Formally, a sequence of $T$ historical frames with $N$ surround-view camera images $\{I_{i, t}\in \mathbb{R}^{H_{i}\times W_{i} \times 3}\}$ are served as input, where $i=1, ..., N$ and $t=1, ..., T$. Besides, sensor intrinsic parameters $\{K_i\}$ together with extrinsic parameters $\{R_i|t_i\}$ for each frame are also provided. Then the ground truth labels describe the voxel states separately, including \textit{occupancy state} and \textit{semantic label}. Three states are considered on the RoboSense dataset, including ``occupied", ``free" and ``unknown". And the semantic label of each voxel can be one of the 3 predefined object categories or an ``unknown" class to indicate general objects. Furthermore, each voxel can be also equipped with extra attributes as outputs, such as instance IDs and motion vectors, which are left as our future work.

To evaluate the quality of predicted occupancy, we measure the whole-scene level voxel segmentation results using $\mathbf{IoU}$ metric for each class. Considering the low-speed driving scenarios, we evaluate the metric under different ranges around the ego vehicle in both 3D and BEV space. Finally, \textbf{mIoU} is obtained through averaging over 4 classes. Moreover, evaluation is only performed on the visible voxels from the camera view.




\section{Experiments}


\begin{table*}[]\footnotesize\centering
\begin{center}
\caption{The details of RoboSense dataset, including the proportion of day/night data among different scenes respectively; The distribution of training/testing/validation sets;  The count of synchronized sequences/frames as well as annotated 3D boxes/trajectories for each scene.}
\label{table_dataset_details}
\vspace{-0.3cm}
\setlength{\tabcolsep}{0.35cm}
\begin{tabular}{c|ccc|ccc|c|c|c|c}
\toprule[1pt]
\multirow{2}{*}{\textbf{Scene-ID}}  & \multicolumn{3}{c|}{\textbf{Distribution}} & \multicolumn{3}{c|}{\textbf{Ratio of Dataset}} & \multirow{2}{*}{\begin{tabular}[c]{@{}c@{}} \textbf{Num of} \\ \textbf{Sequences} \end{tabular}} & \multirow{2}{*}{\begin{tabular}[c]{@{}c@{}}\textbf{Num of} \\ \textbf{Frames} \end{tabular}} & \multirow{2}{*}{\begin{tabular}[c]{@{}c@{}} \textbf{Num of} \\ \textbf{3D Boxes} \end{tabular}} & \multirow{2}{*}{\begin{tabular}[c]{@{}c@{}} \textbf{Num of} \\ \textbf{Trajectories} \end{tabular}} \\ \cline{2-7}
  & \multicolumn{1}{c|}{Day}  & \multicolumn{1}{c|}{Night} & Scene & \multicolumn{1}{c|}{Train} & \multicolumn{1}{c|}{Test} & Val &  &  &  \\ \hline  \hline

S-1 & \multicolumn{1}{c|}{56\%} & \multicolumn{1}{c|}{44\%} & 20\% & \multicolumn{1}{c|}{\multirow{5}{*}{50\%}} & \multicolumn{1}{c|}{\multirow{5}{*}{40\%}} & \multirow{5}{*}{10\%}& 1.5K & 26K & 310K & 36K \\ \cline{1-4}  \cline{8-11} 

S-2 & \multicolumn{1}{c|}{69\%} & \multicolumn{1}{c|}{31\%} & 30\% & \multicolumn{1}{c|}{} & \multicolumn{1}{c|}{} & & 2.3K & 42K & 293K & 37K \\ \cline{1-4} \cline{8-11} 

S-3 & \multicolumn{1}{c|}{71\%} & \multicolumn{1}{c|}{29\%} & 17\% & \multicolumn{1}{c|}{} & \multicolumn{1}{c|}{} &  & 1.2K & 22K & 284K &  64K \\ \cline{1-4}  \cline{8-11} 

S-4 & \multicolumn{1}{c|}{83\%} & \multicolumn{1}{c|}{17\%} & 7\% & \multicolumn{1}{c|}{} & \multicolumn{1}{c|}{} &  & 0.5K & 9K & 144K & 22K \\ \cline{1-4} \cline{8-11} 

S-5 & \multicolumn{1}{c|}{70\%} & \multicolumn{1}{c|}{30\%} & 20\% & \multicolumn{1}{c|}{} & \multicolumn{1}{c|}{} &  & 1.6K & 26K & 297K &  44K \\ \cline{1-11} 

S-6 & \multicolumn{1}{c|}{22\%} & \multicolumn{1}{c|}{78\%} & 6\%  & \multicolumn{1}{c|}{0\%} & \multicolumn{1}{c|}{100\%} & 0\%  & 0.5K & 8K & 88K & 13K \\ \hline
\hline

Total  & \multicolumn{1}{c|}{65\%}  & \multicolumn{1}{c|}{35\%}  & 100\% & \multicolumn{1}{c|}{46\%} & \multicolumn{1}{c|}{44\%} & 10\% & 7.6K & 133K & 1.4M & 216K \\ 
\bottomrule[1pt]
\end{tabular}
\end{center}
\vspace{-0.6cm}
\end{table*}

\subsection{Benchmark Setup}

Our RoboSense dataset contains 7.6K sequences (including 130K annotated frames) of synchronized multi-sensor data, covering 6 main categories (including 22 different locations) of outdoor or semi-closed scenarios (i.e., S1-parks, S2-scenic spots, S3-squares, S4-campuses, S5-sidewalks and S6-streets). \textbf{\textit{To protect the data privacy, we conduct a series of data desensitization measures through masking the human faces and car plates as well as road signs from all sensor data}}. The details of RoboSense dataset composition and partitioning are listed in the Tab.~\ref{table_dataset_details}. The RoboSense dataset is collected under various illumination, traffic flow and weather conditions, to ensure the diversity of static background and movable obstacles, thus meeting the demand of different realistic applications.

RoboSense dataset is divided into three parts with a ratio of 50\%, 40\% and 10\%, for the purpose of training, testing and validation respectively. As for the scene partition, one of the 6 collected scenes (i.e. S-6) is assigned to the testing set exclusively, while the remaining scenes are shared among all splits. Ground truth labels of training and validation sets for corresponding task are provided, together with the synchronized multi-sensor raw data. However, the testing set only provides data. Hence algorithms can merely be submitted to our online benchmark for corresponding task evaluation of testing set.


\subsection{Sensor Specifications}
The detailed specifications of all devices are shown in Tab.~\ref{tab_sensor_specification}.
To cover the areas from near to farther areas, we select Cameras with different focal lengths and Field of View (FOV). Besides, 5 LiDAR sensors are installed in our data collection robot, where the top Hesai Pandar40M is served as autolabeller to provide initial annotations for the splicing points of other LiDARs. 11 Ultrasonics sensors are also installed for freespace detection to ensure safety. All devices are synchronized in time via Network Time Protocol (NTP) before data collection, we utilize a time interval of 100ms as the global timestamp, and match the frame from each device with the nearest timestamp adjacent to the global timestamp. This process ultimately yields synchronized multi-sensor data at a frame rate of 10 FPS.

\begin{table}[t] \footnotesize\centering
\begin{center}
\vspace{-0.2cm}
\caption{Sensor Specifications on RoboSense.} \label{tab_sensor_specification}
  \vspace{-0.2cm}
    \resizebox{1.0\columnwidth}{!}{
    \begin{tabular}{c|c|c}
        \toprule[1pt]
        \textbf{Modality} &  \textbf{Sensor}  & \textbf{Details}
        \\
        \hline
        \hline
                        \multirow{2}{*}{\makecell{\\ Camera}}
        &4 $\times$ Camera &\makecell{RGB, 25Hz, 1920 $\times$ 1080\\ FOV:$[{111.78}^{\circ},{63.16}^{\circ}]$} \\
        \cline{2-3}
            &4 $\times$ Fisheye &\makecell{RGB, 25Hz, 1280 $\times$ 720\\ FOV:$[{180.0}^{\circ},{180.0}^{\circ}]$}\\
        \cline{2-3}
        \hline
            \multirow{3}{*}{\makecell{ \\ \\LiDAR}}
            & Hesai Pandar40M &\makecell{64 beams, 10Hz, 384k pps\\             FOV:$[{360.0}^{\circ}$,${-25}^{\circ}$ to ${15}^{\circ}]$}\\
        \cline{2-3}
            & 3 $\times$ Zvision ML30s &\makecell{40 beams, 10Hz, 720k pps\\ FOV:$[{286.48}^{\circ}$,${-25}^{\circ}$ to ${15}^{\circ}]$}\\
        \cline{2-3}
            & Livox Horizon &\makecell{40 beams, 10Hz, 720k pps\\ FOV:$[{286.48}^{\circ}$,${-25}^{\circ}$ to ${15}^{\circ}]$}\\
        \hline
        \multirow{2}{*}{\makecell{Ultrasonics}}
        &3 $\times$ LRU&\makecell{STP-313, 1m-10m, 40kHz, $\pm$1$mm$}\\
        \cline{2-3}
        &8 $\times$ SRU &\makecell{STP-318, 5cm-200cm, 40kHz, $\pm$1$mm$}\\
        \hline
            \multirow{1}{*}{\makecell{Localization}}
            & GPS \& IMU &\makecell{GPS, IMU, AHRS. ${0.2}^{\circ}$ heading, ${0.1}^{\circ}$ roll/pitch, 20mm, \\
            RTK positioning, 1000Hz update rate}\\
        \bottomrule[1pt]
    \end{tabular}}
\end{center}
\vspace{-0.6cm}
\end{table}

\subsection{Implementation Details}
\begin{figure}[t]
\centering
\setlength{\abovecaptionskip}{-0.cm} 
\includegraphics[width=1.0\columnwidth]{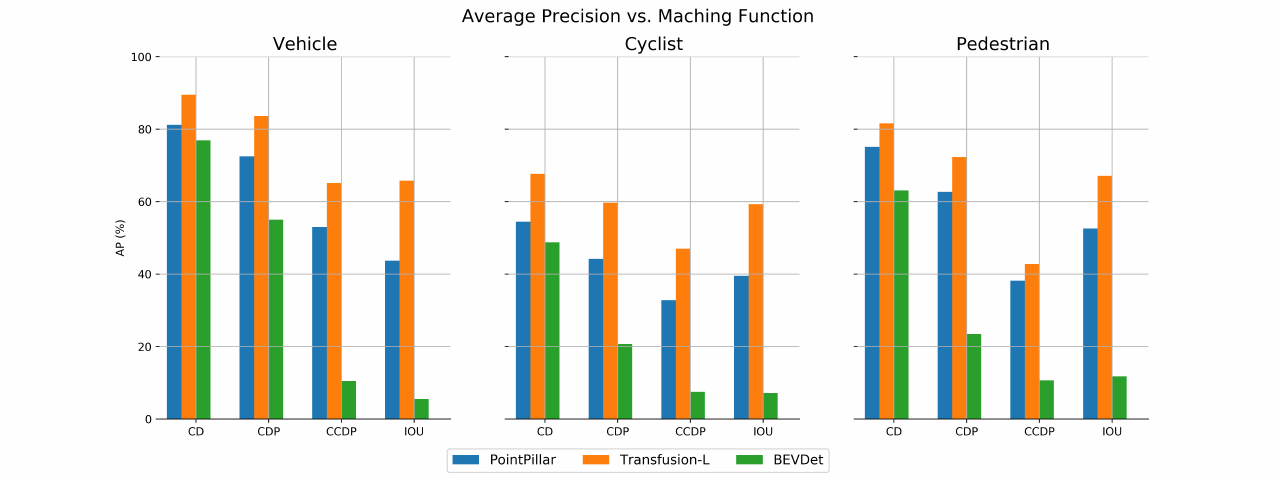}
\caption{Average precision vs. matching function. CD: Center Distance. CDP: Center Distance Proportion. CCDP: Closest-Collision Distance Proportion. IOU: Intersection Over Union. We set IOU of Vehicle, Cyclist and Pedestrian to [0.7, 0.5, 0.5] following KITTI~\cite{geiger2013vision}. CD is set to 2$m$ following nuScenes~\cite{caesar2020nuscenes} and CDP/CCDP=5\% for TP metrics.}
\label{fig:ap_vs_matching}
\vspace{-0.3cm}
\end{figure}

For LiDAR detection task, we set the point range to x$\in$[-45$m$, 45$m$], y$\in$[-45$m$, 45$m$], z$\in$[-1$m$, 4$m$], with a fixed voxel size of 0.16$m$ and 0.05$m$ for pillar-based and voxel-based methods respectively. For Image detection tasks, we use ResNet18~\cite{he2016deep} as backbone network and the input image is resized to $640\times352$. For practical usages, we report performance using our proposed \textit{Closest-Collision Distance Proportion} (CCDP) as matching criterion. Comparisons of different matching functions on average precision are shown in Fig.~\ref{fig:ap_vs_matching}. As expected, when using \textit{Center Distance} (CD) or IOU, objects without distance differentiation can not reflect the model capability of locating closest collision points of nearby obstacles, which is more challenging and essential for low-speed driving scenarios.

\begin{table*}[t]\footnotesize
\begin{center}
\caption{3D Detection results on validation sets of RoboSense using \textit{Center-Point} (CP) distance and \textit{Closest Collision-Point} (CCP) distance as matching criteria respectively where the relative proportion $p$ is set to 5\% (LiDAR) and 10\% (Image).} 
\label{table_comparison_3D_detection}
\vspace{-0.3cm}
\setlength{\tabcolsep}{0.19cm}
\begin{tabular}{c|c|ccc|ccc|ccc}
\toprule[1pt]
\multirow{2}{*}{\textbf{Task}} & \multirow{2}{*}{\textbf{Method}} & \multicolumn{3}{c|}{\textbf{Vehicle@$p$=5\%/10\%}} & \multicolumn{3}{c|}{\textbf{Cyclist@$p$=5\%/10\%}} & \multicolumn{3}{c}{\textbf{Pedestrian@$p$=5\%/10\%}} \\ \cline{3-11}
 &  & \multicolumn{1}{c|}{\textbf{3D AP$\uparrow$} } & \multicolumn{1}{c|}{\textbf{AOS$\uparrow$}} & \textbf{ASE$\downarrow$} & \multicolumn{1}{c|}{\textbf{3D AP$\uparrow$}} & \multicolumn{1}{c|}{\textbf{AOS$\uparrow$}} & \textbf{ASE$\downarrow$} & \multicolumn{1}{c|}{\textbf{3D AP$\uparrow$}} & \multicolumn{1}{c|}{\textbf{AOS$\uparrow$}} & \textbf{ASE$\downarrow$} \\ \hline \hline
 \multirow{4}{*}{\begin{tabular}[c]{@{}c@{}}LiDAR 3D\\ Detection\end{tabular}} & 
  PointPillar~\cite{lang2019pointpillars} & \multicolumn{1}{c|}{72.5/53.0} & \multicolumn{1}{c|}{73.5/61.1} & 20.6/16.1 & \multicolumn{1}{c|}{44.2/32.8} & \multicolumn{1}{c|}{45.4/38.3} & 64.2/54.3 & \multicolumn{1}{c|}{62.7/38.2} & \multicolumn{1}{c|}{45.3/34.1} & 38.3/27.2 \\ \cline{2-11} &
  SECOND~\cite{yan2018second} & \multicolumn{1}{c|}{78.8/63.1} & \multicolumn{1}{c|}{80.2/69.4} & 19.8/15.7 & \multicolumn{1}{c|}{53.8/43.5} & \multicolumn{1}{c|}{57.2/49.9} & 67.7/55.7 & \multicolumn{1}{c|}{70.8/\textbf{47.2}} & \multicolumn{1}{c|}{54.6/43.2} & 40.1/29.3 \\ \cline{2-11} 
 & PVRCNN~\cite{shi2020pv} & \multicolumn{1}{c|}{74.6/57.4} & \multicolumn{1}{c|}{77.4/67.7} & \textbf{16.4}/15.4 & \multicolumn{1}{c|}{53.6/41.4} & \multicolumn{1}{c|}{55.7/50.1} & 62.5/61.9 & \multicolumn{1}{c|}{66.4/39.1} & \multicolumn{1}{c|}{50.1/37.0} & 40.4/25.5 \\ \cline{2-11} 
 & Transfusion-L~\cite{bai2022transfusion} & \multicolumn{1}{c|}{\textbf{83.6/65.1}} & \multicolumn{1}{c|}{\textbf{84.5/73.8}}  & \multicolumn{1}{c|}{19.7/16.0} & \multicolumn{1}{c|}{\textbf{59.7/47.0}} &  \multicolumn{1}{c|}{\textbf{78.0/70.8}} & \multicolumn{1}{c|}{82.1/72.9} & \multicolumn{1}{c|}{\textbf{72.3}/42.8} & \multicolumn{1}{c|}{\textbf{60.5/48.7}} & 45.1/37.4  \\ 

 \hline
\multirow{4}{*}{\begin{tabular}[c]{@{}c@{}}Multi-view 3D\\ Detection\end{tabular}} 
 & BEVDet~\cite{huang2021bevdet} & \multicolumn{1}{c|}{76.2/30.2} & \multicolumn{1}{c|}{40.4/25.9} & 17.3/11.2 & \multicolumn{1}{c|}{42.3/25.7} & \multicolumn{1}{c|}{36.1/30.2} & 56.5/42.1 & \multicolumn{1}{c|}{47.4/28.5} & \multicolumn{1}{c|}{48.6/36.5} & 30.2/18.8 \\ \cline{2-11} 
 & BEVDet4D~\cite{huang2022bevdet4d} & \multicolumn{1}{c|}{77.2/31.1} & \multicolumn{1}{c|}{41.1/26.4} & \multicolumn{1}{c|}{16.8/10.8} & \multicolumn{1}{c|}{42.0/24.8} & \multicolumn{1}{c|}{33.9/27.7} & 55.3/\textbf{41.2} & \multicolumn{1}{c|}{48.1/29.3} & \multicolumn{1}{c|}{46.6/37.6} & \textbf{27.5}/21.3 \\ \cline{2-11} 
& BEVDepth~\cite{li2023bevdepth} & \multicolumn{1}{c|}{77.8/31.3} & \multicolumn{1}{c|}{40.9/26.3} & \multicolumn{1}{c|}{16.7/10.7} & \multicolumn{1}{c|}{43.3/27.0} & \multicolumn{1}{c|}{34.9/30.2} & \textbf{52.2}/46.6 & \multicolumn{1}{c|}{50.1/31.3} & \multicolumn{1}{c|}{46.7/37.9} & 28.0/21.4 \\ \cline{2-11} 
  & BEVFormer~\cite{li2022bevformer} & \multicolumn{1}{c|}{78.2/32.0} & \multicolumn{1}{c|}{41.6/26.7} & 16.5/\textbf{10.6} & \multicolumn{1}{c|}{44.1/27.6} & \multicolumn{1}{c|}{34.9/30.5} & 51.3/44.3 & \multicolumn{1}{c|}{50.2/32.3} & \multicolumn{1}{c|}{46.3/38.0} & 28.1/\textbf{17.9} \\ 
 \bottomrule[1pt]
\end{tabular}
\end{center}
\vspace{-0.5cm}
\end{table*}

\begin{table*}[t]\footnotesize
\begin{center}
\caption{Study of different sensor layouts for perception tasks (3D detection and MOT) on validation sets of RoboSense under different ranges (m). AB3DMOT~\cite{2020AB3DMOT} is adopted as 3D MOT baseline. C: Camera, F: Fisheye, L: LiDAR, V: View} \label{tab_sensor_layouts}
\vspace{-0.3cm}
\setlength{\tabcolsep}{0.23cm}
\begin{tabular}{c|c|c|cccc|cccc}
\toprule[1pt]
\multirow{3}{*}{\textbf{Task}} & \multirow{3}{*}{\textbf{Detector}} & \multirow{3}{*}{\textbf{Layouts}} & \multicolumn{4}{c|}{\textbf{Detection}} & \multicolumn{4}{c}{\textbf{Tracking}} \\ \cline{4-11} 
 & & & \multicolumn{1}{c|}{\multirow{2}{*}{\textbf{Metric}}} & \multicolumn{3}{c|}{\textbf{Range($m$)}} & \multicolumn{1}{c|}{\multirow{2}{*}{
\begin{tabular}[c]{@{}c@{}} \textbf{sAMOTA$\uparrow$} \end{tabular}}}
  & \multicolumn{1}{c|}{\multirow{2}{*}{\begin{tabular}[c]{@{}c@{}} \textbf{AMOTP$\uparrow$} \end{tabular}}} & \multicolumn{1}{c|}{\multirow{2}{*}{\textbf{MT$\uparrow$}}} & \multirow{2}{*}{\textbf{ML$\downarrow$}} \\ \cline{5-7}
 & &  & \multicolumn{1}{c|}{} & \multicolumn{1}{c|}{{[}\cellcolor[HTML]{DADADA}0, 5{]}} & \multicolumn{1}{c|}{{[}5, 10{]}} & {[}10, 30{]} & \multicolumn{1}{c|}{} & \multicolumn{1}{c|}{} & \multicolumn{1}{c|}{} &  \\ \hline \hline

\multirow{6}{*}{\begin{tabular}[c]{@{}c@{}} Multi-view \\ 3D Perception\end{tabular}} & \multirow{6}{*}{\begin{tabular}[c]{@{}c@{}}BEVDepth~\cite{li2023bevdepth}\end{tabular}} &\multirow{2}{*}{\begin{tabular}[c]{@{}c@{}}4C\end{tabular}} & \multicolumn{1}{c|}{3D AP} & \multicolumn{1}{c|}{\cellcolor[HTML]{DADADA}54.9/16.0} & \multicolumn{1}{c|}{60.1/18.3} & 53.7/33.1 & \multicolumn{1}{c|}{\multirow{2}{*}{44.03}} & \multicolumn{1}{c|}{\multirow{2}{*}{29.95}} & \multicolumn{1}{c|}{\multirow{2}{*}{20.23}} & \multirow{2}{*}{54.01} \\ \cline{4-7}
 & & & \multicolumn{1}{c|}{AOS} & \multicolumn{1}{c|}{\cellcolor[HTML]{DADADA}44.8/19.7} & \multicolumn{1}{c|}{37.0/18.8} & 34.5/26.9 & \multicolumn{1}{c|}{} & \multicolumn{1}{c|}{} & \multicolumn{1}{c|}{} &  \\ \cline{3-11}
 
& & \multirow{2}{*}{\begin{tabular}[c]{@{}c@{}}4F\end{tabular}} & \multicolumn{1}{c|}{3D AP} & \multicolumn{1}{c|}{\cellcolor[HTML]{DADADA}61.1/16.9} & \multicolumn{1}{c|}{70.6/19.9} & 50.8/29.0 & \multicolumn{1}{c|}{\multirow{2}{*}{39.56}} & \multicolumn{1}{c|}{\multirow{2}{*}{27.10}} & \multicolumn{1}{c|}{\multirow{2}{*}{18.02}} & \multirow{2}{*}{61.74} \\ \cline{4-7}
 & & & \multicolumn{1}{c|}{AOS} & \multicolumn{1}{c|}{\cellcolor[HTML]{DADADA}58.7/27.5} & \multicolumn{1}{c|}{41.3/23.5} & 36.1/27.4 & \multicolumn{1}{c|}{} & \multicolumn{1}{c|}{} & \multicolumn{1}{c|}{} &  \\ \cline{3-11}
 
& & \multirow{2}{*}{\begin{tabular}[c]{@{}c@{}}4C + 4F\end{tabular}} & \multicolumn{1}{c|}{3D AP} & \multicolumn{1}{c|}{\cellcolor[HTML]{DADADA}\textbf{68.9}/20.5} & \multicolumn{1}{c|}{\textbf{75.2}/22.9} & 64.2/38.6 & \multicolumn{1}{c|}{\multirow{2}{*}{\textbf{51.16}}} & \multicolumn{1}{c|}{\multirow{2}{*}{35.68}} & \multicolumn{1}{c|}{\multirow{2}{*}{25.21}} & \multirow{2}{*}{48.07} \\ \cline{4-7}
 &  & & \multicolumn{1}{c|}{AOS} & \multicolumn{1}{c|}{\cellcolor[HTML]{DADADA}53.9/24.4} & \multicolumn{1}{c|}{43.1/22.5} & 39.6/30.9 & \multicolumn{1}{c|}{} & \multicolumn{1}{c|}{} & \multicolumn{1}{c|}{} &  \\ \hline
 
 \multirow{2}{*}{\begin{tabular}[c]{@{}c@{}} LiDAR \\3D Perception \end{tabular}} & \multirow{2}{*}{\begin{tabular}[c]{@{}c@{}} PointPillar~\cite{lang2019pointpillars} \end{tabular}} & \multirow{2}{*}{\begin{tabular}[c]{@{}c@{}}4L\end{tabular}} & \multicolumn{1}{c|}{3D AP} & \multicolumn{1}{c|}{\cellcolor[HTML]{DADADA}59.2/19.3} & \multicolumn{1}{c|}{73.1/42.0} & \textbf{71.0/65.4} & \multicolumn{1}{c|}{\multirow{2}{*}{44.77}} & \multicolumn{1}{c|}{\multirow{2}{*}{33.65}} & \multicolumn{1}{c|}{\multirow{2}{*}{25.04}} & \multirow{2}{*}{54.08} \\ \cline{4-7}
 & & & \multicolumn{1}{c|}{AOS} & \multicolumn{1}{c|}{\cellcolor[HTML]{DADADA}46.5/19.2} & \multicolumn{1}{c|}{67.2/47.5} & 69.0/65.7 & \multicolumn{1}{c|}{} & \multicolumn{1}{c|}{} & \multicolumn{1}{c|}{} &  \\ \hline

 \multirow{2}{*}{\begin{tabular}[c]{@{}c@{}} Multi-modal \\ 3D Perception \end{tabular}} & \multirow{2}{*}{\begin{tabular}[c]{@{}c@{}} BEVDepth~\cite{li2023bevdepth} \\ + Pointpillar~\cite{lang2019pointpillars} \end{tabular}} & \multirow{2}{*}{\begin{tabular}[c]{@{}c@{}}8V + 4L\end{tabular}} & \multicolumn{1}{c|}{3D AP} & \multicolumn{1}{c|}{\cellcolor[HTML]{DADADA}61.3/\textbf{36.9}} & \multicolumn{1}{c|}{61.3/\textbf{54.6}} & 54.4/52.6 & \multicolumn{1}{c|}{\multirow{2}{*}{43.32}} & \multicolumn{1}{c|}{\multirow{2}{*}{\textbf{43.18}}} & \multicolumn{1}{c|}{\multirow{2}{*}{\textbf{34.74}}} & \multirow{2}{*}{\textbf{40.82}} \\ \cline{4-7}
 & & & \multicolumn{1}{c|}{AOS} & \multicolumn{1}{c|}{\cellcolor[HTML]{DADADA}\textbf{64.8/49.6}} & \multicolumn{1}{c|}{\textbf{78.7/75.0}} & \textbf{79.4/78.4} & \multicolumn{1}{c|}{} & \multicolumn{1}{c|}{} & \multicolumn{1}{c|}{} &  \\ 
 
\bottomrule[1pt]
\end{tabular}
\end{center}
\vspace{-0.4cm}
\end{table*}

\subsection{Baselines: Perception}
\vspace{-0.5cm}
\noindent
\subsubsection{LiDAR 3D Detection} 
To demonstrate the performance of advanced 3D detectors on LiDAR-only detection track of our RoboSense benchmark, we implement several popular CNN-based methods with different fashions, including Pointpillar~\cite{lang2019pointpillars} (Pillar-based), SECOND~\cite{yan2018second} (Voxel-based), and PV-RCNN~\cite{shi2020pv} (Two-stage Point-Voxel based). Besides, Transformer-based method such as Transfusion-L~\cite{bai2022transfusion} is also implemented for architecture comparison. Pointpillar as the most efficient method above is adopted as our baseline for LiDAR 3D detection task.


\noindent
\subsubsection{Multi-View 3D Detection}
Current works of multi-view 3D detection can be divided into two mainstreams, namely LSS~\cite{philion2020lift} based and Transformer based. To examine the effectiveness of image-only multi-view 3D detection models, we select the widely-used method BEVDet~\cite{huang2021bevdet} as our LSS-based baseline on image 3D detection track of RoboSense, and re-implement several extended versions such as BEVDet4D~\cite{huang2022bevdet4d} which takes advantage of history temporal clues, and BEVDepth~\cite{li2023bevdepth} which adopts an additional branch for depth prediction under point supervision. Besides, BEVFormer~\cite{li2022bevformer} as a Transformer-based representative work is also included. 



\noindent
\subsubsection{Multiple Object Tracking}
We follow the ``Tracking-by-Detection" paradigm using 3D detection results from Camera or LiDAR data as input respectively, and present several baselines for multiple 3D object tracking task. Specifically, 3D boxes detected from surround-view images by BEVDepth~\cite{li2023bevdepth} and splicing pointclouds by Pointpillar~\cite{lang2019pointpillars} are provided separately. And the tracking approach AB3DMOT described in~\cite{2020AB3DMOT} is picked to serve as the baseline of multiple object tracker in the 3D space. Then the same objects across different sensors are associated with unique track IDs to form global trajectories in the past.


\subsection{Baselines: Prediction}
\vspace{-0.5cm}

\noindent
\subsubsection{Motion Prediction} 
Traditional motion prediction methods utilize perception ground truth (i.e., history trajectories of agents and HDmap) as input, which lacks of uncertainty modeling in practical applications. In this paper, we implement several vision-based end-to-end methods for joint perception and motion prediction on RoboSense benchmark, including ViP3D~\cite{gu2023vip3d} and PnPNet~\cite{liang2020pnpnet}. For comparisons, we also report the motion prediction results of assuming agents surrounding the ego-vehicle with constant positions or velocities respectively, thus to reflect the diversity and difficulty of our dataset on prediction task.



\noindent
\subsubsection{Occupancy Prediction}
We extend a BEV 3D detection model - BEVDepth~\cite{li2023bevdepth} to the 3D occupancy prediction task, which is then adopted as our baseline for the visual occupancy prediction task. Concretely, we replace the original detection decoders with the occupancy reconstruction layers while maintaining the BEV feature encoders. ResNet18~\cite{he2016deep} pretrained on FCOS3D~\cite{wang2021fcos3d} is employed as image backbone for visual feature extraction.

\subsection{Results and Analysis}


\vspace{-0.3cm}
\noindent
\subsubsection{Perception Results} 
\textbf{3D Object Detection.} The 3D detection results based on multi-view images and splicing point clouds are shown in Tab.~\ref{table_comparison_3D_detection}. As for LiDAR 3D detection, Transfusion-L~\cite{bai2022transfusion} achieves the leading performance owing to the advanced transformer architecture. In terms of multi-view 3D detection, BEVDet4D~\cite{huang2022bevdet4d} and BEVDepth~\cite{li2023bevdepth} obtain significant improvement than BEVDet~\cite{huang2021bevdet} through involving temporal clues and adopting an additional depth branch respectively. Besides, BEVFormer~\cite{li2022bevformer} also achieves competitive results by introducing a query-based attention mechanism. Generally, LiDAR-based 3D detector can generate high-quality detection results than vision-based methods. However, vision-based methods are capable of detecting various ranges of objects with more sensors (Fisheye or Camera). Note that two different matching criteria are both considered for TP calculation, namely Center-Point (CP) distance and Closest Collision-Point (CCP) distance. It can be observed that the CCP localization performance is obviously lower than the CP localization (\textit{i.e.} 18.5\% 3D AP drop of Transfusion-L for Vehicle class and 29.5\% 3D AP drop for Pedestrian class. For navigation safety, the CCP localization is more important for near-field egocentric perception in crowded social scenarios.

\textbf{Performance with Different Sensor-layouts.} To evaluate the performance of different sensor layouts under various ranges, we conduct extensive comparisons as shown in Tab.~\ref{tab_sensor_layouts}. As for visual perception, 4C layout achieves better AP than 4F layout in farther areas (i.e., 10-30$m$), while 4F layout is good at detecting near-field targets within 10$m$. Through combining these two layouts, better performance can be achieved across different ranges. LiDAR 3D detector exhibits an obvious advantage over visual detectors especially in CCP and farther object localization, while the performance of near-field objects within 5$m$ is inferior (19.3\% vs. 20.5\%). Moreover, we implement multi-modal 3D perception (8V+4L) through late-fusion strategy. Specifically, 3D detection results from multi-view 3D detector and LiDAR 3D detector are adopted for post-processing. And we can observe that the CCP-based 3D AP of objects within 5$m$ is remarkably boosted from 20.5\% to 36.9\%. And the AOS metric is also increased consistently.

\textbf{Multiple Object Tracking.} Regarding to the MOT task in Tab.~\ref{tab_sensor_layouts}, AB3DMOT~\cite{2020AB3DMOT} is adopted as baseline tracker in 3D space, which mitigates the impact of object occlusions existing in 2D image, especially for crowded scenarios. Through introducing more sensors (4C + 4F), vision-based methods can also achieve competitive tracking performance with LiDAR-based methods, even better in sAMOTA metric (51.16 vs. 44.77). With the multi-modal input, AMOTP, MT and ML performance can be further improved as expected. However, although equipped with multi-modal and multi-sensor data as input, \textbf{the perception performance is still inferior especially in near-field} 
(\textit{i.e.} 36.9\% CCP-based 3D AP within 5\textit{m}), revealing the deficiencies of current perception methods in handling the obstacles in near ranges. \textbf{The main reason may be the frequent truncation and occlusion caused by a large view occupation of near obstacles}, which showcases the great challenge and importance of our proposed benchmark for the development of egocentric perceptual frameworks related to navigation in crowded and unstructured environments.


\begin{table}[t]
\centering
\caption{Motion forecasting results on validation sets of RoboSense. $*$ and $\dag$  
 indicate utilizing GroundTruth 3D boxes and detection results from PointPillar~\cite{lang2019pointpillars} as input respectively with constant positions or velocities for comparisons.}\label{tab_motion_forecasting}
\centering
\vspace{-0.3cm}
\resizebox{1.0\columnwidth}{!}{
\begin{tabular}{c|cccc}
\toprule[1pt]
\textbf{Method} & \cellcolor[HTML]{DADADA} \textbf{minADE (\textit{m}) $\downarrow$} & \textbf{minFDE (\textit{m})$\downarrow$} & \textbf{MR$\downarrow$} & \textbf{EPA$\uparrow$}  \\ \hline\hline
\begin{tabular}[c]{@{}c@{}}Constant Pos.*\end{tabular} & \cellcolor[HTML]{DADADA}2.42 & 3.01 & 0.319 & 0.680  \\
\begin{tabular}[c]{@{}c@{}}Constant Vel.*\end{tabular} & \cellcolor[HTML]{DADADA}1.59 & 3.54 & 0.219 & 0.780  \\
\begin{tabular}[c]{@{}c@{}}Constant Pos.$^{\dag}$\end{tabular} & \cellcolor[HTML]{DADADA}1.52 & 1.95 & 0.267 & 0.243  \\
\begin{tabular}[c]{@{}c@{}}Random Vel.$^{\dag}$\end{tabular} & \cellcolor[HTML]{DADADA}2.56 & 3.85 & 0.872 & 0.029  \\
\begin{tabular}[c]{@{}c@{}}ViP3D~\cite{gu2023vip3d}\end{tabular} & \cellcolor[HTML]{DADADA}1.31 & 1.55 & 0.196 & 0.283\\ 
\begin{tabular}[c]{@{}c@{}}PnPNet~\cite{liang2020pnpnet}\end{tabular} & \cellcolor[HTML]{DADADA}0.89 & 1.12 & 0.172 & 0.313 \\ 
\bottomrule[1pt]
\end{tabular}}
\vspace{-0.2cm}
\end{table}

\begin{table}[t]
\centering
\caption{Occupancy prediction results on validation sets of RoboSense using 4F sensors as input. “mIoU-3D” and “mIoU-BEV” indicate the standard mIoU metric calculated in 3D space and BEV respectively without considering the ground voxels.}\label{tab_occ_prediction}
\vspace{-0.3cm}
\setlength{\tabcolsep}{1cm}
\resizebox{1.0\columnwidth}{!}{
\begin{tabular}{c|cc}
\toprule[1pt]
\textbf{Range(m)} & \cellcolor[HTML]{DADADA}\textbf{mIoU-3D$\uparrow$} & \textbf{mIoU-BEV$\uparrow$}  \\ \hline\hline
\begin{tabular}[c]{@{}c@{}}[0, 12.8]\end{tabular} & \cellcolor[HTML]{DADADA} 24.6 & 29.7  \\ 
\hline
\begin{tabular}[c]{@{}c@{}}[0, 2]\end{tabular}  & \cellcolor[HTML]{DADADA} 39.6 & 48.2 \\ 
\begin{tabular}[c]{@{}c@{}}[2, 5]\end{tabular} & \cellcolor[HTML]{DADADA} 30.7 & 36.7 \\ 
\begin{tabular}[c]{@{}c@{}}[5, 12.8]\end{tabular}  & \cellcolor[HTML]{DADADA} 16.1 & 19.7  \\ 
\bottomrule[1pt]
\end{tabular}}
\vspace{-0.3cm}
\end{table}
 
\vspace{-0.4cm}
\noindent
\subsubsection{Prediction Results} 
Motion forecasting of surrounding agents as well as occupancy state descriptions around the ego-vehicle are two crucial prediction tasks in the research field of autonomous driving, which have been extensively explored in urban and highway scenarios for autonomous cars. 

\textbf{Motion Prediction.} As shown in Tab.~\ref{tab_motion_forecasting}, either visual end-to-end methods~\cite{gu2023vip3d} or LiDAR-based end-to-end methods~\cite{liang2020pnpnet} are all supported for validation on our RoboSense. PnPNet~\cite{liang2020pnpnet} with LiDAR points as input can produce less prediction errors and better EPA than ViP3D~\cite{gu2023vip3d}, both of which remarkably outperform two baseline settings of modeling agents with constant positions or velocities. 

\textbf{Occupancy Prediction.} As shown in Tab.~\ref{tab_occ_prediction}, we use 4F sensor data as input and report the performance of mIOU metric in both 3D and BEV space under various ranges respectively. Note that the metric is calculated without considering states of the ground voxels, leading to lower performance in either 3D or BEV space. As expected, the performance evaluated within 2\textit{m} is better than farther areas.

\section{Conclusion}

To foster the research of egocentric perceptual framework tailored to various types of autonomous agents navigating in crowded and unstructured environments, \textbf{RoboSense}, a real-world and multi-modal dataset is collected in complex social scenarios with varying and uncontrolled environmental conditions and dynamical elements. It consists of 7.6K scenes manually selected from different locations, with 1.4M 3D Boxes and 216K trajectories annotated in total on 133K synchronous frames. Besides, occupancy descriptions are also provided to facilitate the surrounding context comprehension. In the future works, more tasks and associated benchmarks, such as motion planning, will be expanded for end-to-end autonomous navigating application, and explore the additional benefits that joint optimization can bring to the modular training.


{
    \small
    \bibliographystyle{ieeenat_fullname}
    \bibliography{main}
}

\clearpage
\maketitlesupplementary
\appendix
\setcounter{page}{1}
\setcounter{section}{0}
\setcounter{figure}{0}
\setcounter{equation}{0}
\renewcommand{\thefigure}{A\arabic{figure}}

\section{Coordinates Transformation}
\subsection{LiDAR$\Leftrightarrow$Ego-Vehicle}
\noindent
\quad\textbf{LiDAR to Ego-Vehicle:}
$(x_v, y_v, z_v)$ represents a three-dimensional coordinate point in Ego-Vehicle Coordinate System. The transformation from the coordinates $(x_v, y_v, z_v)$ in the Ego-Vehicle Coordinate System to $(x_l, y_l, z_l)$ in the LiDAR Coordinate System is calculated as follows:
\begin{equation}
\left(\begin{array}{l}
x_l \\
y_l \\
z_l \\
1
\end{array}\right)=\left[\begin{array}{cc}
R_L^{3 \times 3} & T_L^{3 \times 1} \\
0 & 1
\end{array}\right]\left(\begin{array}{c}
x_v \\
y_v \\
z_v \\
1
\label{eq:lidar2ego}
\end{array}\right)
\end{equation}
where $R_L \in \mathbb{R}^{3 \times 3}$ and $T_L \in \mathbb{R}^{3 \times 1}$ represent the rotation and translation from the Ego-Vehicle Coordinate System to the LiDAR Coordinate System, respectively.

\noindent
\quad\textbf{Ego-Vehicle to LiDAR:}
The transformation from Ego-Vehicle Coordinate System to LiDAR Coordinate System is the inverse transformation of Eq.\eqref{eq:lidar2ego}.

\subsection{LiDAR$\Leftrightarrow$Camera}

\noindent
\quad\textbf{LiDAR to Camera:}
Regardless of whether it is a fisheye or a pinhole camera, the coordinate transformation formula from the LiDAR Coordinate System to the Camera Coordinate System is the same and is given as follows:
\begin{equation}
\left(\begin{array}{c}
x_c \\
y_c \\
z_c \\
1
\end{array}\right)=\left[\begin{array}{cc}
R_C^{3 \times 3} & T_C^{3 \times 1} \\
0 & 1
\end{array}\right]\left(\begin{array}{c}
x_l \\
y_l \\
z_l \\
1
\label{eq:lidar2camera}
\end{array}\right)
\end{equation}
where $(x_c, y_c, z_c)$ represents a three-dimensional coordinate point in the Camera Coordinate System. $R_C \in \mathbb{R}^{3 \times 3}$ and $T_C \in \mathbb{R}^{3 \times 1}$ represent the rotation and translation from the LiDAR Coordinate System to the Camera Coordinate System, respectively.

\noindent
\quad\textbf{Camera to LiDAR:}
The transformation from Camera Coordinate System to LiDAR Coordinate System is the inverse transformation of Eq.\eqref{eq:lidar2camera}.

\subsection{Camera$\Leftrightarrow$Pixel}

\noindent
\quad\textbf{Camera to Pixel:}
The projection formulas of different types of cameras are different in the RoboSense dataset, the projection formula of a pinhole camera is as follows:
\begin{equation}
z_c\left(\begin{array}{l}
u \\
v \\
1
\end{array}\right)=K^{3 \times 3}\left(\begin{array}{l}
x_c \\
y_c \\
z_c \\
\end{array}\right), K^{3 \times 3}=\left[\begin{array}{ccc}
f_x & -1 & u_0 \\
0 & f_y & v_0 \\
0 & 0 & 1
\label{eq:camera2pixel}
\end{array}\right]
\end{equation}
$(u, v)$ is pixel coordinate, $K \in \mathbb{R}^{3 \times 1}$ represents the camera intrinsic parameters, $(f_x, f_y)$ represents the focal lengths of the camera, and $(u_0, v_0)$ indicates the displacement of the camera's optical center from the origin of the Pixel Coordinate System. The projection formula from camera coordinate to pixel coordinate of the fisheye camera is very different, the camera projection process refers to the projection formula of Omnidirectional Camera (OCam) in ~\cite{scaramuzza2006flexible}.

\noindent
\quad\textbf{Pixel to Camera:} 
The transformation from Pixel Coordinate System to Camera Coordinate System in a pinhole camera model requires the inverse of Eq.\eqref{eq:camera2pixel}. Since this is a 2D to 3D transformation, it is necessary to first determine the magnitude of $z_c$. The projection formula from pixel coordinate to camera coordinate of the fisheye camera refers to the projection formula of Omnidirectional Camera (OCam) in ~\cite{scaramuzza2006flexible}.

\subsection{Ego-Vehicle$\Leftrightarrow$Global} 

\noindent
\quad\textbf{Ego-Vehicle to Global:}
$R_G \in \mathbb{R}^{3 \times 3}$ and $T_G \in \mathbb{R}^{3 \times 1}$ represent the transformation matrices of the vehicle's orientation and position in the Global Coordinate System, respectively. The transformation formula for converting the coordinates $(x_v, y_v, z_v)$ in the Ego-Vehicle Coordinate System to $(x_g, y_g, z_g)$ in the Global Coordinate System is as follows:
\begin{equation}
\left(\begin{array}{l}
x_g \\
y_g \\
z_g \\
1
\end{array}\right)=\left[\begin{array}{cc}
R_G^{3 \times 3} & T_G^{3 \times 1} \\
0 & 1
\end{array}\right]\left(\begin{array}{c}
x_v \\
y_v \\
z_v \\
1
\label{eq:ego2global}
\end{array}\right)
\end{equation}

\noindent
\quad\textbf{Global to Ego-Vehicle:}
The transformation from Global Coordinate System to Ego-Vehicle Coordinate System is the inverse transformation of Eq.\eqref{eq:ego2global}.

\begin{table*}[]\centering \small
\caption{The Number and proportion of 3D Boxes from all sensors (Global Scenes) and Livox LiDAR (Local Scenes) per category under different ranges (m) respectively.}
\vspace{-0.3cm}
\setlength{\tabcolsep}{0.26cm}
\begin{tabular}{c|ccc|ccc|ccc|c}
\toprule[1pt]
\multirow{2}{*}{\textbf{Global/Local}} &  \multicolumn{3}{c|}{\textbf{Vehicle}}  & \multicolumn{3}{c|}{\textbf{Cyclist}}  & \multicolumn{3}{c|}{\textbf{Pedestrian}}  & \multirow{2}{*}{\textbf{Total}} \\ \cline{2-10}

& \multicolumn{1}{c|}{\textbf{[0 - 10]} } & \multicolumn{1}{c|}{\textbf{[10 - 30]} } & \multicolumn{1}{c|}{\textbf{[30 - ]} }  & \multicolumn{1}{c|}{\textbf{[0 - 10]} } & \multicolumn{1}{c|}{\textbf{[10 - 30]} } & \multicolumn{1}{c|}{\textbf{[30 - ]} }  & \multicolumn{1}{c|}{\textbf{[0 - 10]} } & \multicolumn{1}{c|}{\textbf{[10 - 30]} } & \multicolumn{1}{c|}{\textbf{[30 - ]}}  &  \\ \hline\hline

\multirow{3}{*}{\begin{tabular}[c]{@{}c@{}}Global\\ (Hesai LiDAR)\end{tabular}} & \multicolumn{1}{c|}{165K} & \multicolumn{1}{c|}{402K} & 343K & \multicolumn{1}{c|}{23K} & \multicolumn{1}{c|}{38K} & 15K  & \multicolumn{1}{c|}{187K} & \multicolumn{1}{c|}{163K} & \multicolumn{1}{c|}{51K} & \multirow{2}{*}{1.4M} \\ \cline{2-10} 
 & \multicolumn{3}{c|}{910K} & \multicolumn{3}{c|}{76K} & \multicolumn{3}{c|}{401K}  &  \\ \cline{2-11} 
 & \multicolumn{3}{c|}{65.00\%} & \multicolumn{3}{c|}{5.42\%} & \multicolumn{3}{c|}{28.64\% } & 100\% \\ \hline\hline
 
\multirow{3}{*}{\begin{tabular}[c]{@{}c@{}}Local\\ (Livox LiDAR)\end{tabular}} & \multicolumn{1}{c|}{150K} & \multicolumn{1}{c|}{282K} & 133K & \multicolumn{1}{c|}{20K} & \multicolumn{1}{c|}{28K} & 7K & \multicolumn{1}{c|}{163K} & \multicolumn{1}{c|}{103K} & 21K & \multirow{2}{*}{907K} \\ \cline{2-10} 
 & \multicolumn{3}{c|}{565K} & \multicolumn{3}{c|}{55K} & \multicolumn{3}{c|}{287K}  &  \\ \cline{2-11} 
 & \multicolumn{3}{c|}{40.36\%} & \multicolumn{3}{c|}{3.93\%} & \multicolumn{3}{c|}{20.50\%} & 64.79\% \\
\bottomrule[1pt]
\end{tabular}
\label{tab:dataset_comparison}
\end{table*}

\begin{figure*}
\centering
\includegraphics[width=14cm]{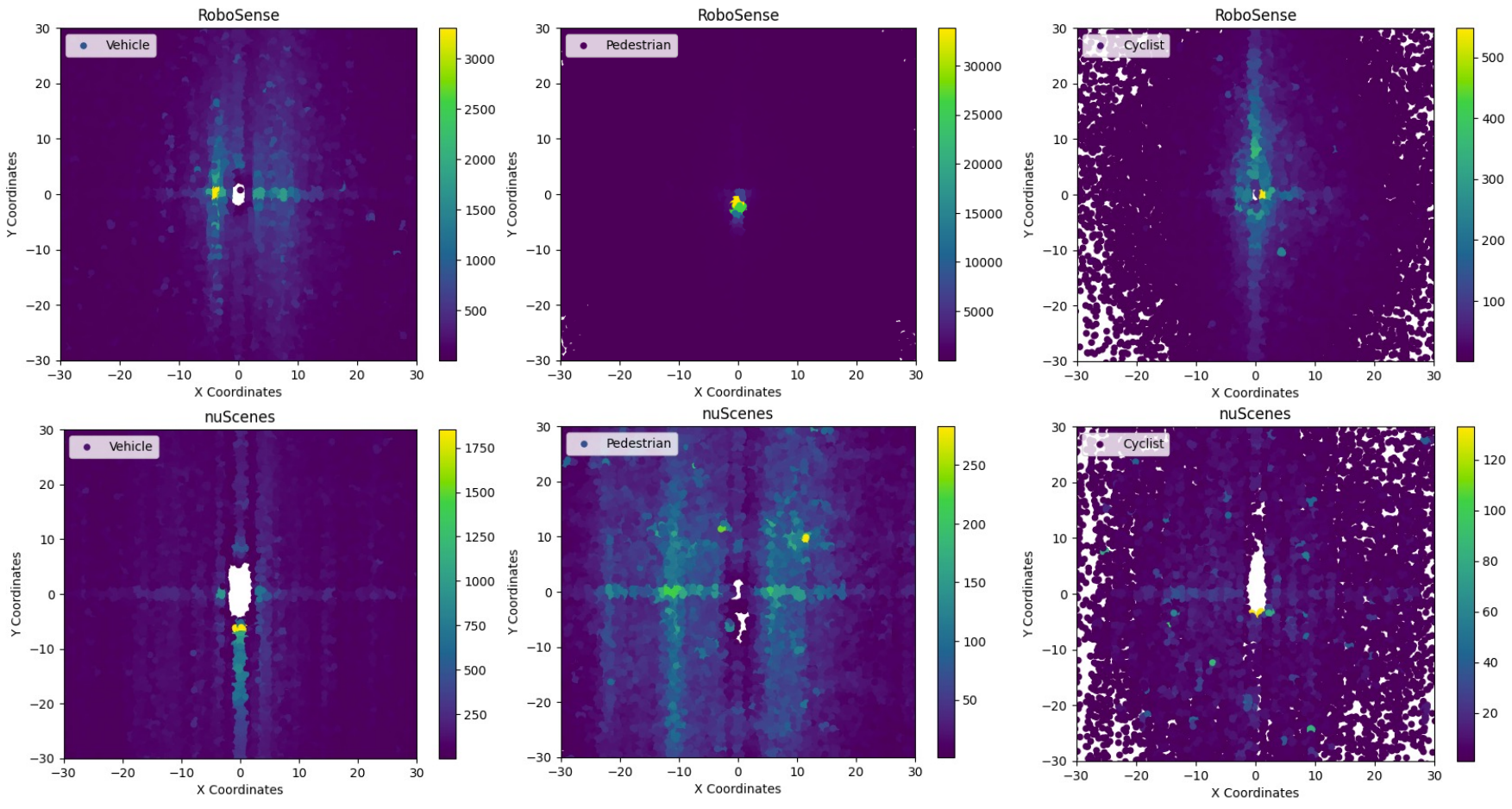}
\caption{Comparison of annotated object distribution of different classes between RoboSense and nuScenes datasets.}
\label{fig:dataset_comparison2}
\end{figure*}

\section{More Details of RoboSense}

\subsection{Annotation Statistics}
We present more statistics on the annotations of RoboSense as shown in Tab.~\ref{tab:dataset_comparison}. It can be observed that our RoboSense dataset contains approximately 1.4M annotated objects, with vehicles and pedestrians comprising the majority, while cyclists are lesser. The distribution of objects is relatively uniform in terms of distance. Additionally, due to the smaller coverage area of Livox pointclouds (Local view) compared to Hesai pointclouds (Global view), the number of annotated objects in the Livox pointclouds is only 64.79\% of that in the Hesai pointclouds. In Fig.~\ref{fig:dataset_comparison2}, we further compare the distribution of annotated objects between our Robosense dataset and nuScenes dataset. It is obvious that our Robosense dataset contains significantly more annotated objects of vehicles, pedestrians, and cyclists classes respectively, which tend to be closer to the ego robot.

\subsection{3D Object Label Generation}
\label{sec:label_gen}
To generate high-quality 3D object annotations, we design a three-stage 3D object generation pipeline for different sensors covering various ranges. First, a pre-trained LiDAR detection model (i.e., \cite{lang2019pointpillars}) of high precision is adopted to produce 3D objects on the full $360\degree$ view using high-quality Pandar64 points as input. Then expert annotators are required to refine the initial 3D boxes continuously throughout the whole sequences in each scene, based on splicing pointclouds which are obtained by aligning 4 vehicle-side LiDARs to the Ego-Vehicle coordinate through affine transformation. Besides, annotators need to supplement surrounding 3D boxes in a near range which are not scanned by the top Hesai LiDAR or fail to be detected owing to high occlusion and truncation. Last but not least, invalid 3D annotations should be excluded for target LiDAR coordinate and Camera coordinate respectively, where the annotated objects are not covered in the corresponding sensor data. Through multiple validation steps, highly accurate annotations can be achieved in both near and far ranges. We also release intermediate Pandar64 points for research usages.

\begin{table*}[t]\footnotesize
\begin{center}
\caption{3D Detection results of different modalities on validation sets of RoboSense using \textit{IoU} as matching criteria.} \label{tab_3d_detection}
\label{table_comparison_3D_detection_iou}
\vspace{-0.3cm}
\setlength{\tabcolsep}{0.25cm}
\begin{tabular}{c|c|ccc|ccc|ccc}
\toprule[1pt]
\multirow{2}{*}{\textbf{Task}} & \multirow{2}{*}{\textbf{Method}} & \multicolumn{3}{c|}{\textbf{Vehicle@IoU=0.7/0.3}} & \multicolumn{3}{c|}{\textbf{Cyclist@IoU=0.5/0.3}} & \multicolumn{3}{c}{\textbf{Pedestrian@IoU=0.5/0.3}} \\ \cline{3-11}
 &  & \multicolumn{1}{c|}{\textbf{3D AP$\uparrow$} } & \multicolumn{1}{c|}{\textbf{AOS$\uparrow$}} & \textbf{ASE$\downarrow$} & \multicolumn{1}{c|}{\textbf{3D AP$\uparrow$}} & \multicolumn{1}{c|}{\textbf{AOS$\uparrow$}} & \textbf{ASE$\downarrow$} & \multicolumn{1}{c|}{\textbf{3D AP$\uparrow$}} & \multicolumn{1}{c|}{\textbf{AOS$\uparrow$}} & \textbf{ASE$\downarrow$} \\ \hline \hline

 \multirow{4}{*}{\begin{tabular}[c]{@{}c@{}}LiDAR 3D\\ Detection\end{tabular}} & 
  PointPillar~\cite{lang2019pointpillars} & \multicolumn{1}{c|}{43.7} & \multicolumn{1}{c|}{45.5} & 13.3 & \multicolumn{1}{c|}{39.5} & \multicolumn{1}{c|}{39.6} & 69.2 & \multicolumn{1}{c|}{52.6} & \multicolumn{1}{c|}{36.6} & 34.9 \\ \cline{2-11} &
  SECOND~\cite{yan2018second} & \multicolumn{1}{c|}{55.8} & \multicolumn{1}{c|}{59.8} & 17.2 & \multicolumn{1}{c|}{52.3} & \multicolumn{1}{c|}{53.3} & 65.9 & \multicolumn{1}{c|}{61.7} & \multicolumn{1}{c|}{46.9} & 37.5 \\ \cline{2-11} 
 & PVRCNN~\cite{shi2020pv} & \multicolumn{1}{c|}{53.5} & \multicolumn{1}{c|}{57.9} & 16.9 & \multicolumn{1}{c|}{53.0} & \multicolumn{1}{c|}{50.7} & 55.9 & \multicolumn{1}{c|}{58.9} & \multicolumn{1}{c|}{43.4} & 38.4 \\ \cline{2-11} 
 & Transfusion-L~\cite{bai2022transfusion} & \multicolumn{1}{c|}{\textbf{65.8}} & \multicolumn{1}{c|}{\textbf{66.3}} & \multicolumn{1}{c|}{17.3} & \multicolumn{1}{c|}{\textbf{59.3}} & \multicolumn{1}{c|}{\textbf{71.0}} & \multicolumn{1}{c|}{78.5} & \multicolumn{1}{c|}{\textbf{67.1}} & \multicolumn{1}{c|}{\textbf{56.0}} & 42.7  \\ 

 \hline
\multirow{4}{*}{\begin{tabular}[c]{@{}c@{}}Multi-view 3D\\ Detection\end{tabular}} 
 & BEVDet~\cite{huang2021bevdet} & \multicolumn{1}{c|}{32.1} & \multicolumn{1}{c|}{21.8} & 10.4 & \multicolumn{1}{c|}{19.9} & \multicolumn{1}{c|}{21.2} & 36.8 & \multicolumn{1}{c|}{25.9} & \multicolumn{1}{c|}{29.7} & 20.3 \\ \cline{2-11} 
 & BEVDet4D~\cite{huang2022bevdet4d} & \multicolumn{1}{c|}{33.5} & \multicolumn{1}{c|}{22.8} & 10.4 & \multicolumn{1}{c|}{20.1} & \multicolumn{1}{c|}{21.1} & 36.7 & \multicolumn{1}{c|}{26.2} & \multicolumn{1}{c|}{28.3} & \textbf{17.7} \\ \cline{2-11} 
  & BEVDepth~\cite{li2023bevdepth} & \multicolumn{1}{c|}{33.4} & \multicolumn{1}{c|}{22.8} & \textbf{10.2} & \multicolumn{1}{c|}{22.6} & \multicolumn{1}{c|}{22.2} & 41.6 & \multicolumn{1}{c|}{27.7} & \multicolumn{1}{c|}{28.1} & 17.9 \\ \cline{2-11} 
  & BEVFormer~\cite{li2022bevformer} & \multicolumn{1}{c|}{33.6} & \multicolumn{1}{c|}{23.0} & 10.3 & \multicolumn{1}{c|}{23.4} & \multicolumn{1}{c|}{22.1} & \textbf{35.3} & \multicolumn{1}{c|}{28.0} & \multicolumn{1}{c|}{29.5} & 17.8 \\ 
 \bottomrule[1pt]
\end{tabular}
\end{center}
\vspace{-0.5cm}
\end{table*}

\begin{figure}[t]
\centering
\includegraphics[width=1.0\columnwidth]{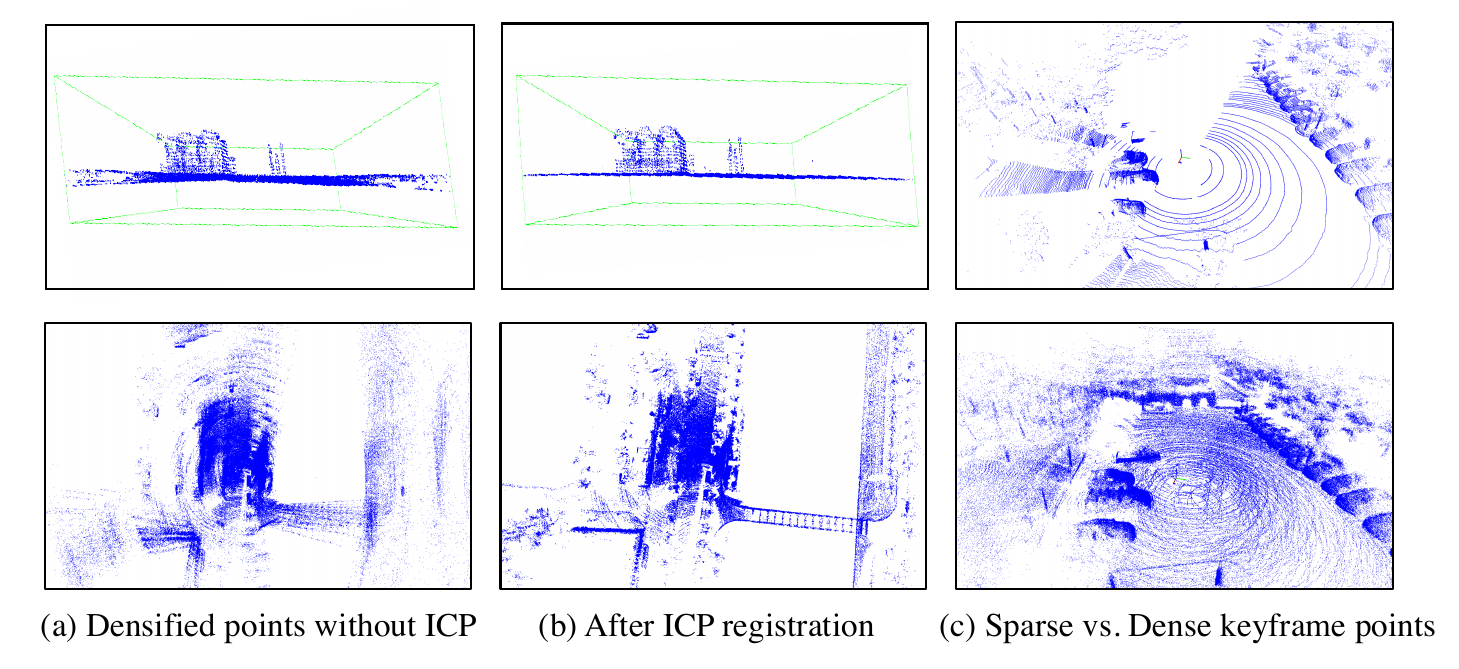}
\caption{Illustration of ICP and points densified process.}
\label{fig:icp_points}
\vspace{-0.5cm}
\end{figure}

\subsection{Occupancy Label Preprocess}
\label{sec:occ_gen}
Occupancy label generation can be primarily divided into two parts: pointclouds densification and occupancy label determination. Unlike existing counterpart~\cite{tang2024sparseocc} which only utilizes the sparse keyframe LiDAR points, multi-frame aggregation operation is found to be indispensable for dense occupancy generation. For dynamic objects, the extracted dynamic points of neighboring frames are subsequently concatenating for each object along the corresponding trajectory respectively, thus achieving the pointclouds densification. For static scenes, coordinate transformation is performed from the ego-vehicle coordinate to the global coordinate across time using ego-pose information, and then simply aggregate all static points on the ego-vehicle coordinate of current keyframe through concatenation.

Notably, owing to the complex driving scenarios with uneven ground and rapid pose changes especially when turning directions to avoid obstacles during data collection, pose drifts are observed in the IMU data. Therefore, the temporal aggregation results of pointclouds are inferior with misaligned horizon and ego-motion blur as shown in Fig.~\ref{fig:icp_points}. To relieve these issues, ICP (Interative-Closed-Point)~\cite{sharp2002icp} is conducted additionally for static scene points registeration before multi-frame aggregation. Finally, densified pointclouds for a single frame can be obtained by fusing the static scenes with the dynamic objects.

Given dense points of a specific scene, we label all voxels within a fixed range by a resolution of $0.5m\times0.5m$, based on the height of majority points inside each voxel. If the height is larger than a threshold $\sigma$, the voxel state is set to ``occupied", otherwise ``free". Moreover, considering the occlusion and truncation situations, some occupied voxels are not scanned by LiDAR beams and camera views actually. Hence we set part of voxels to ``unknown" state which are invisible from both the LiDAR and camera views through tracing the casting ray.


\subsection{Metric Comparison}
In addition to the evaluation of 3D detection results with the proposed matching criteria (\textit{Center-Point} distance and \textit{Closest Collision-Point} distance), we also provide the corresponding evaluation results using the traditional 3D \textit{IOU} (Intersection-Over-Union) matching criteria for comparison, as shown in Tab.~\ref{table_comparison_3D_detection_iou}.
It is obvious that without distance differentiation, the evaluation results of 3D AP for both LiDAR-based and Camera-based methods are all in a low level, which can not reflect the objective performance and fail to satisfy the practical application requirements of the detection model. However, the proposed matching criterion is designed to measure the  locating capability of closest collision points of nearby obstacles, which is more challenging and essential for low-speed driving scenarios.

\subsection{Scene Distribution} 

\begin{figure}[t]
\centering
\includegraphics[width=0.95\columnwidth]{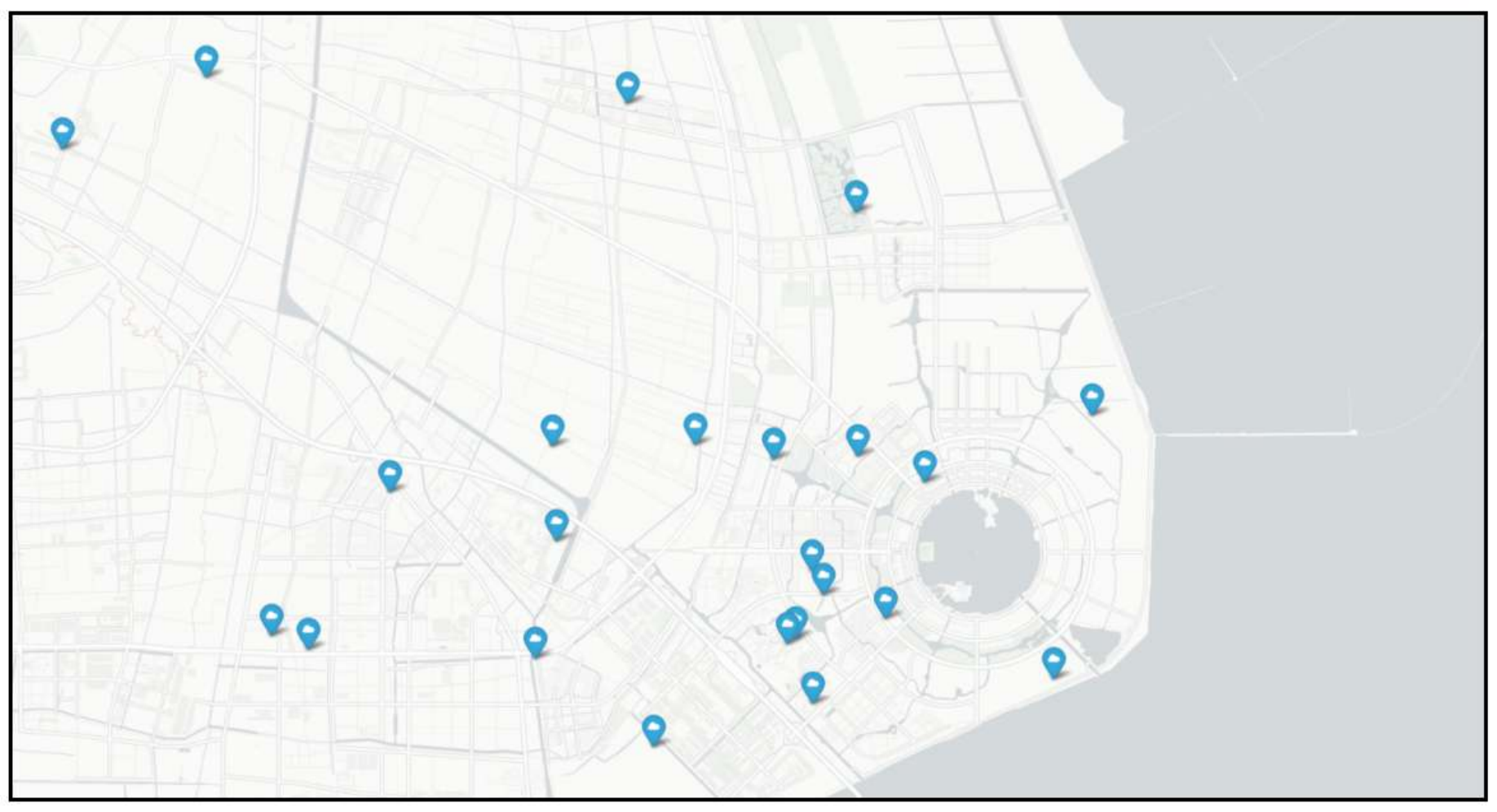}
\caption{Distribution of data collection scenarios in RoboSense dataset in Google Map.}
\label{fig:scene_distribution}
\vspace{-0.3cm}
\end{figure}

\begin{figure*}
\centering
\setlength{\abovecaptionskip}{-0.cm} 
\includegraphics[width=16cm]{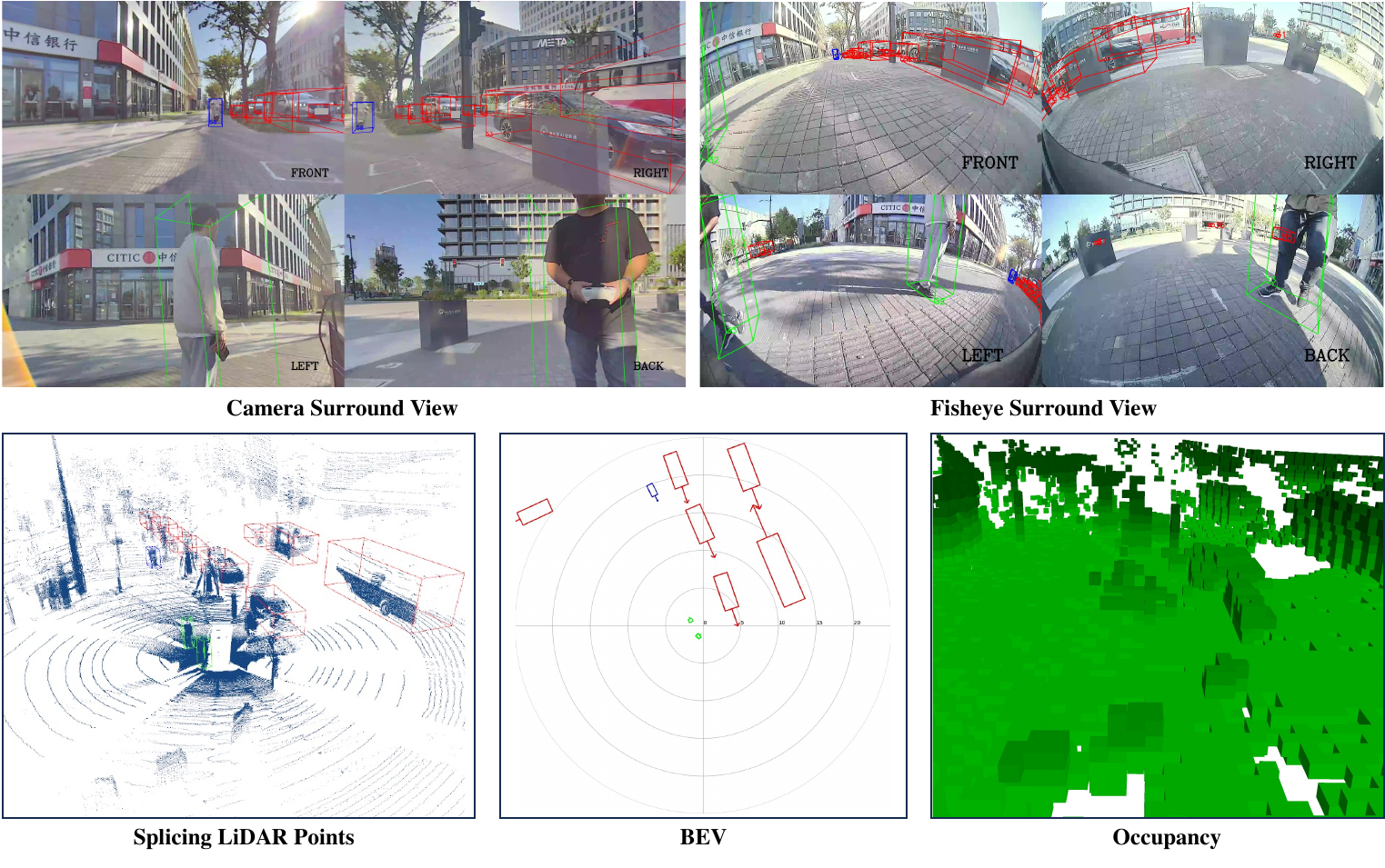}
\caption{The illustration of S1-parks in Sequence-4906 at the 3-rd frame.}
\label{fig:vis1}
\end{figure*}

\begin{figure*}
\centering
\setlength{\abovecaptionskip}{-0.cm} 
\includegraphics[width=16cm]{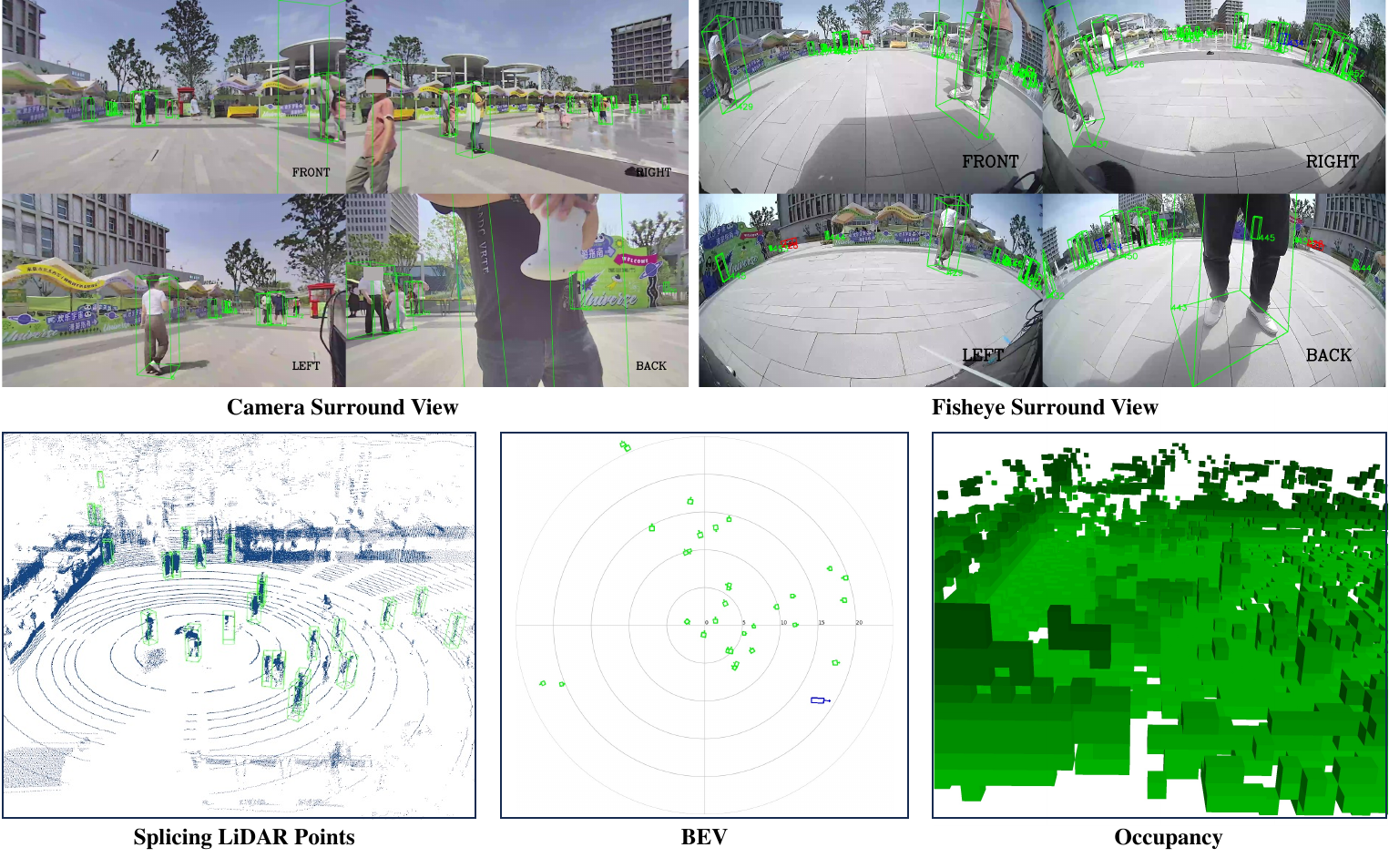}
\caption{The illustration of S2-scenic spots in Sequence-1491 at the 13-th frame.}
\label{fig:vis2}
\end{figure*}

\begin{figure*}
\centering
\setlength{\abovecaptionskip}{-0.cm} 
\includegraphics[width=16cm]{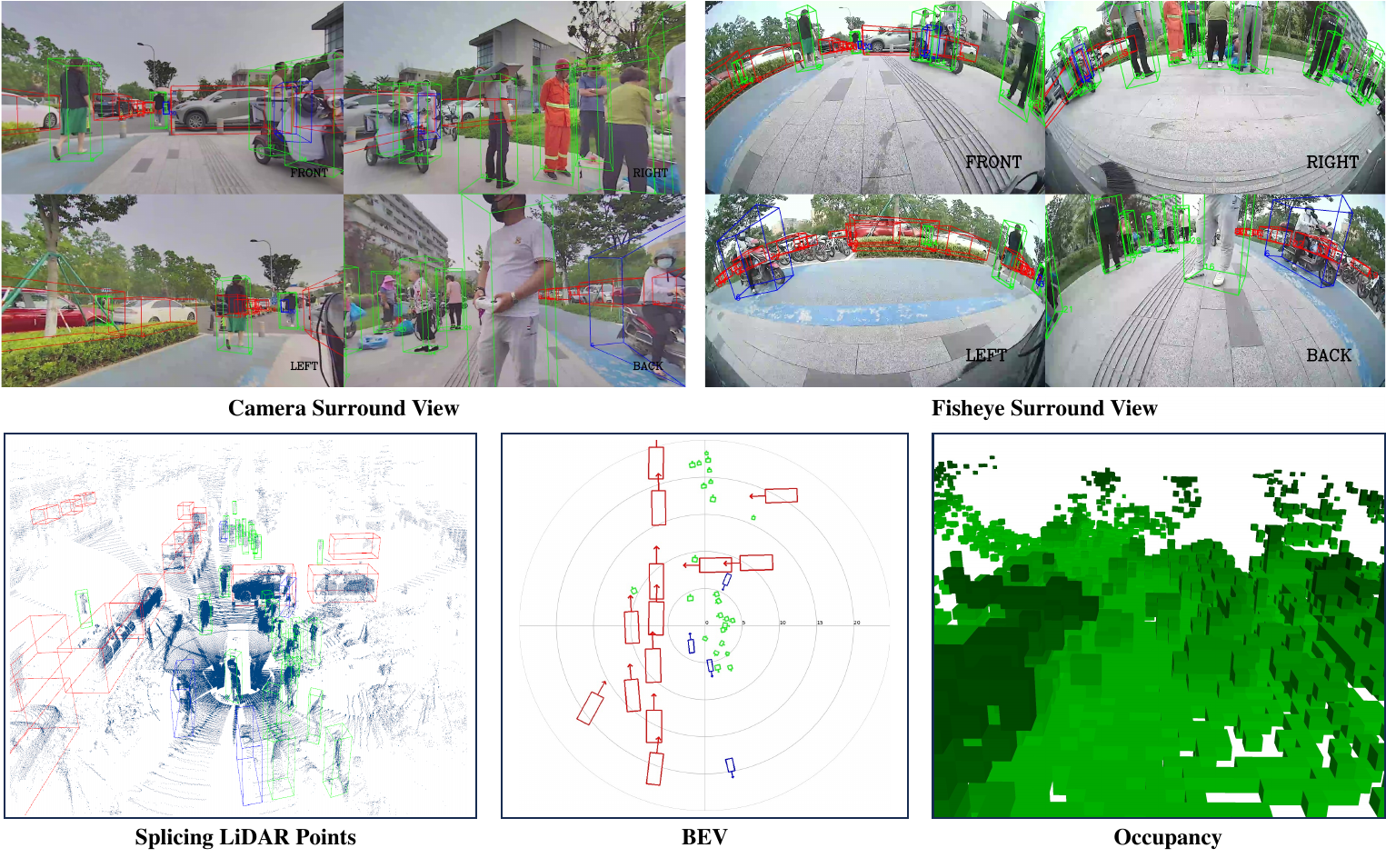} 
\caption{The illustration of S3-squares in Sequence-396 at the 2-nd frame.}
\label{fig:vis3}
\end{figure*}

\begin{figure*}
\centering
\setlength{\abovecaptionskip}{-0.cm} 
\includegraphics[width=16cm]{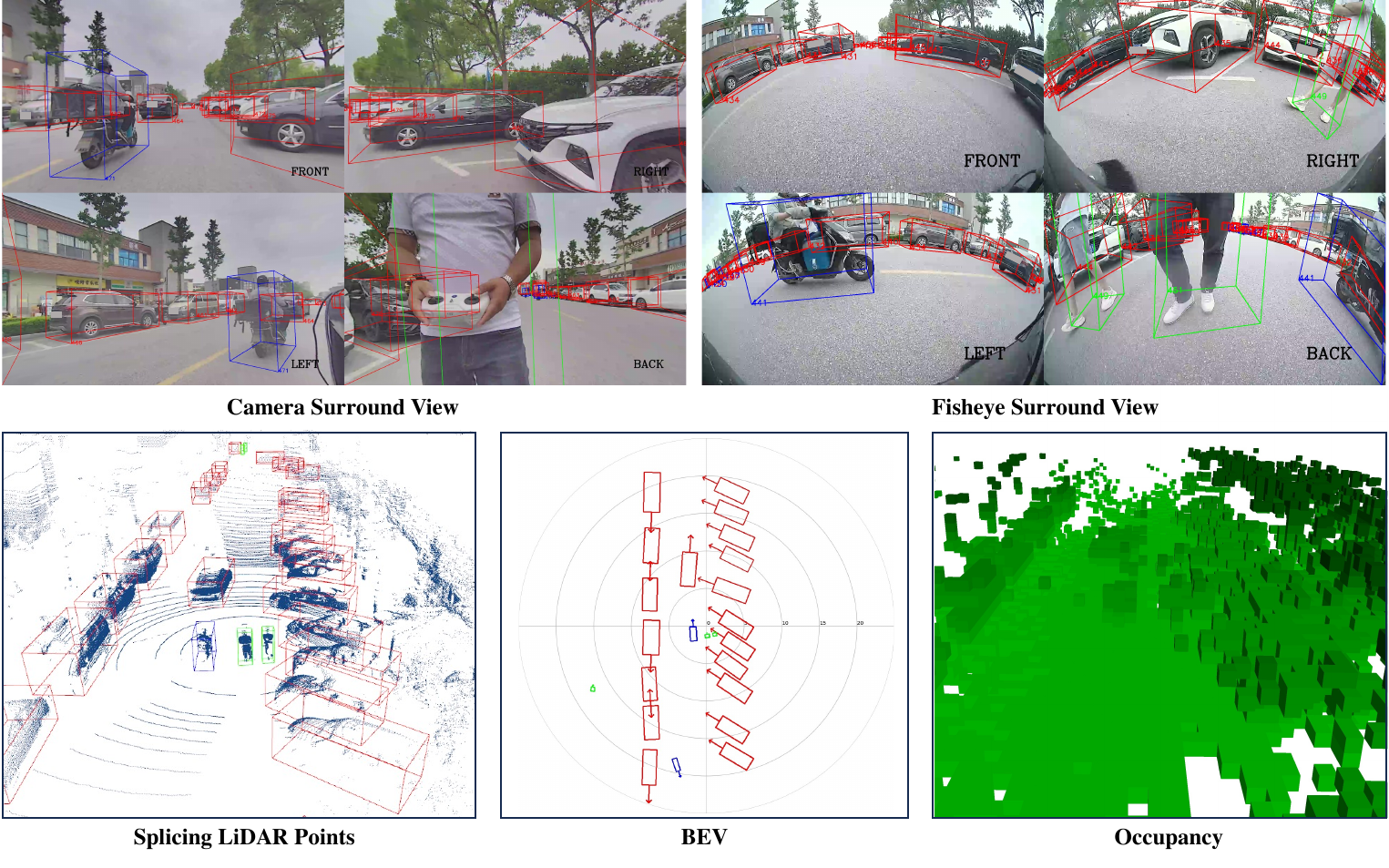}
\caption{The illustration of S4-campuses in Sequence-2257 at the 16-th frame.}
\label{fig:vis4}
\end{figure*}

\begin{figure*}
\centering
\setlength{\abovecaptionskip}{-0.cm} 
\includegraphics[width=16cm]{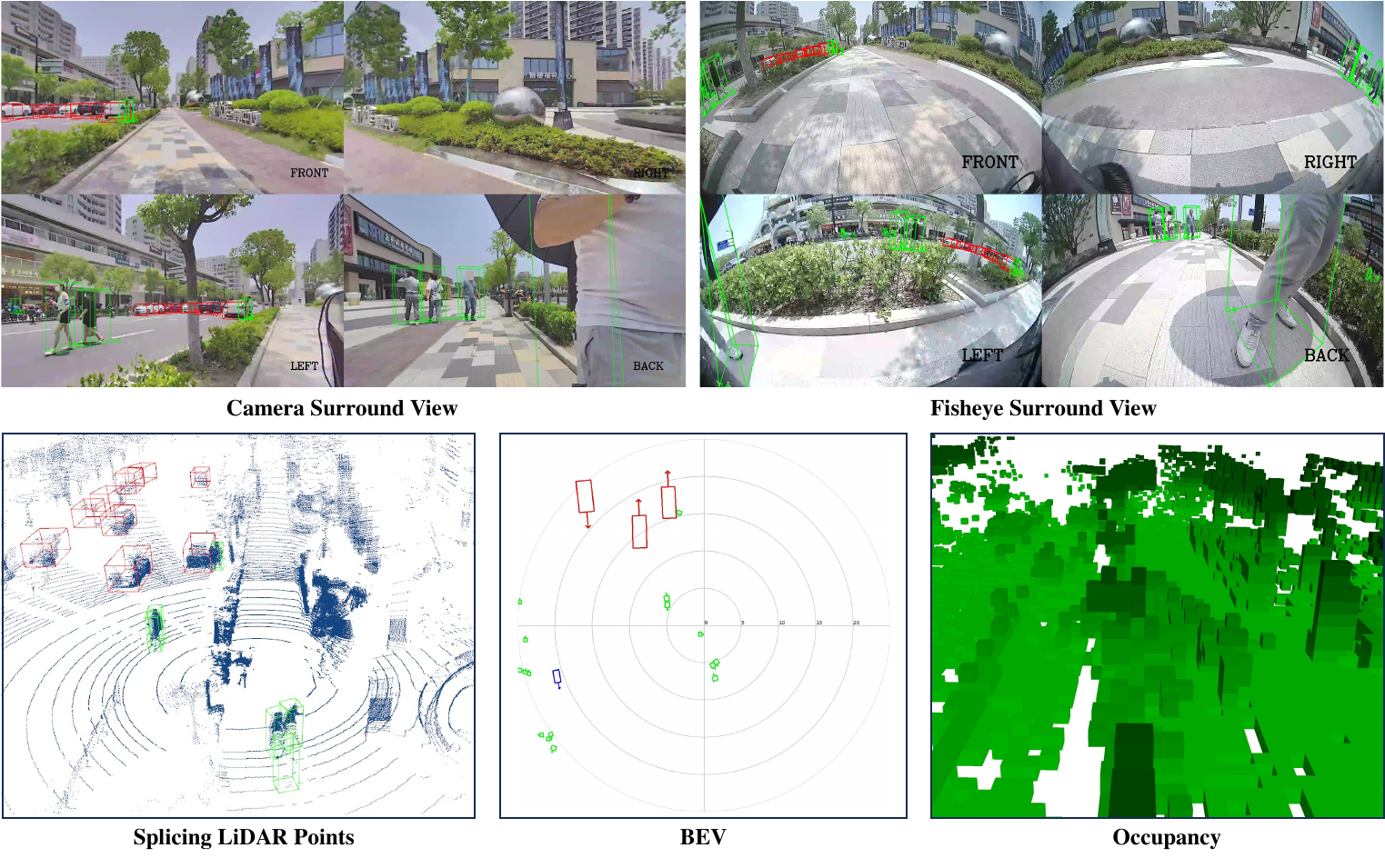}
\caption{The illustration of S5-sidewalks in Sequence-2990 at the 10-th frame.}
\label{fig:vis5}
\end{figure*}

\begin{figure*}
\centering
\setlength{\abovecaptionskip}{-0.cm} 
\includegraphics[width=16cm]{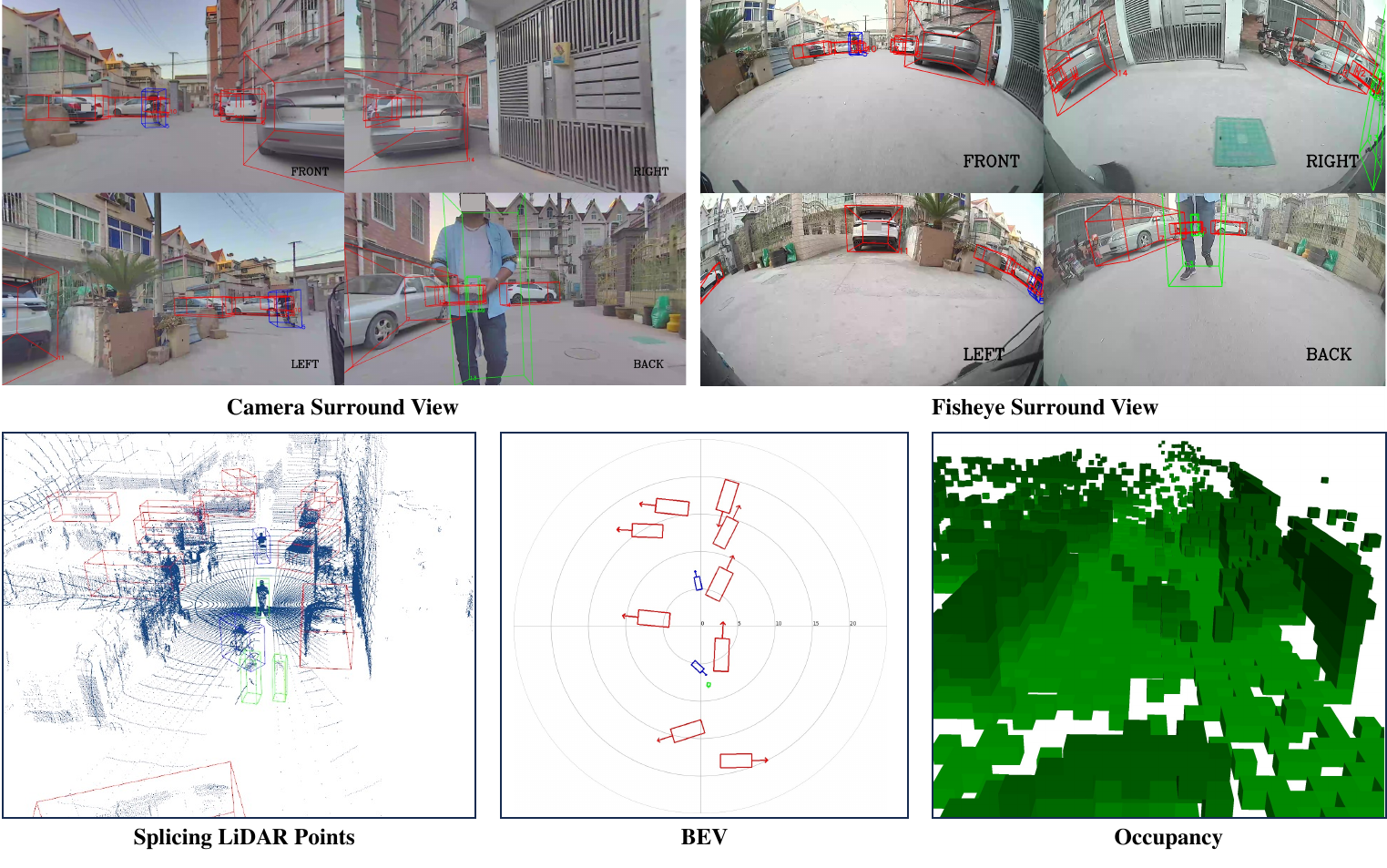}
\caption{The illustration of S6-streets in Sequence-7018 at the 2-nd frame.}
\label{fig:vis6}
\end{figure*}

Our RoboSense dataset contains 7.6K sequences, covering 6 main categories (including 22 different locations) of outdoor or semi-closed scenarios (i.e., S1-parks, S2-scenic spots, S3-squares, S4-campuses and S5-sidewalks or S6-streets). Fig.~\ref{fig:scene_distribution} illustrates the scene distributions of our collected data constructed for RoboSense dataset, which are surrounding Dishui Lake in Shanghai, China, with several markers drew in Google Map indicating the main locations performed data collection. Besides, the illustrations for each representative scenario among the collected locations are shown in Fig.~\ref{fig:vis1}-\ref{fig:vis6} respectively.



\end{document}